\documentclass[Afour,sageh,times]{sagej}

\usepackage{moreverb,url}

\usepackage[colorlinks,bookmarksopen,bookmarksnumbered,citecolor=red,urlcolor=red]{hyperref}

\newcommand\BibTeX{{\rmfamily B\kern-.05em \textsc{i\kern-.025em b}\kern-.08em
T\kern-.1667em\lower.7ex\hbox{E}\kern-.125emX}}

\def\volumeyear{2016}
\usepackage{graphicx} 
\usepackage{amsmath} 
\usepackage{amsfonts}

\usepackage{siunitx}
\usepackage[capitalize]{cleveref}

\usepackage{array}
\usepackage{ulem}

\usepackage{xspace}

\usepackage{multirow}

\usepackage{url}

\usepackage[colorinlistoftodos, disable]{todonotes}

\begin{document}

\runninghead{Agarwal, Mirzaee, Sun, and Yuan}
\title{A Modularized Design Approach for Gelsight Family of Vision-based Tactile Sensors}

\author{Arpit Agarwal\affilnum{1,${*}$}, Mohammad Amin Mirzaee\affilnum{2,${*}$}, Xiping Sun\affilnum{2} and Wenzhen Yuan\affilnum{2}}

\affiliation{\affilnum{1}Carnegie Mellon University, USA\\
\affilnum{2}University of Illinois Urbana-Champaign, USA\\
\affilnum{${*}$}These authors contributed equally to this work.
}

\corrauth{Wenzhen Yuan, University of Illinois Urbana-Champaign, USA
\email{yuanwz@illinois.edu}
}

\begin{abstract}

GelSight family of vision-based tactile sensors has proven to be effective for multiple robot perception and manipulation tasks. These sensors are based on an internal optical system and an embedded camera to capture the deformation of the soft sensor surface, inferring the high-resolution geometry of the objects in contact. However, customizing the sensors for different robot hands requires a tedious trial-and-error process to re-design the optical system. In this paper, we formulate the GelSight sensor design process as a systematic and objective-driven design problem and perform the design optimization with a physically accurate optical simulation. The method is based on modularizing and parameterizing the sensor's optical components and designing four generalizable objective functions to evaluate the sensor. We implement the method with an interactive and easy-to-use toolbox called OptiSense Studio. With the toolbox, non-sensor experts can quickly optimize their sensor design in both forward and inverse ways following our predefined modules and steps. We demonstrate our system with four different GelSight sensors by quickly optimizing their initial design in simulation and transferring it to the real sensors. 
\end{abstract}


\maketitle

\section{Introduction}

\begin{figure*}[!t]
    \centering
    \includegraphics[width=\linewidth]{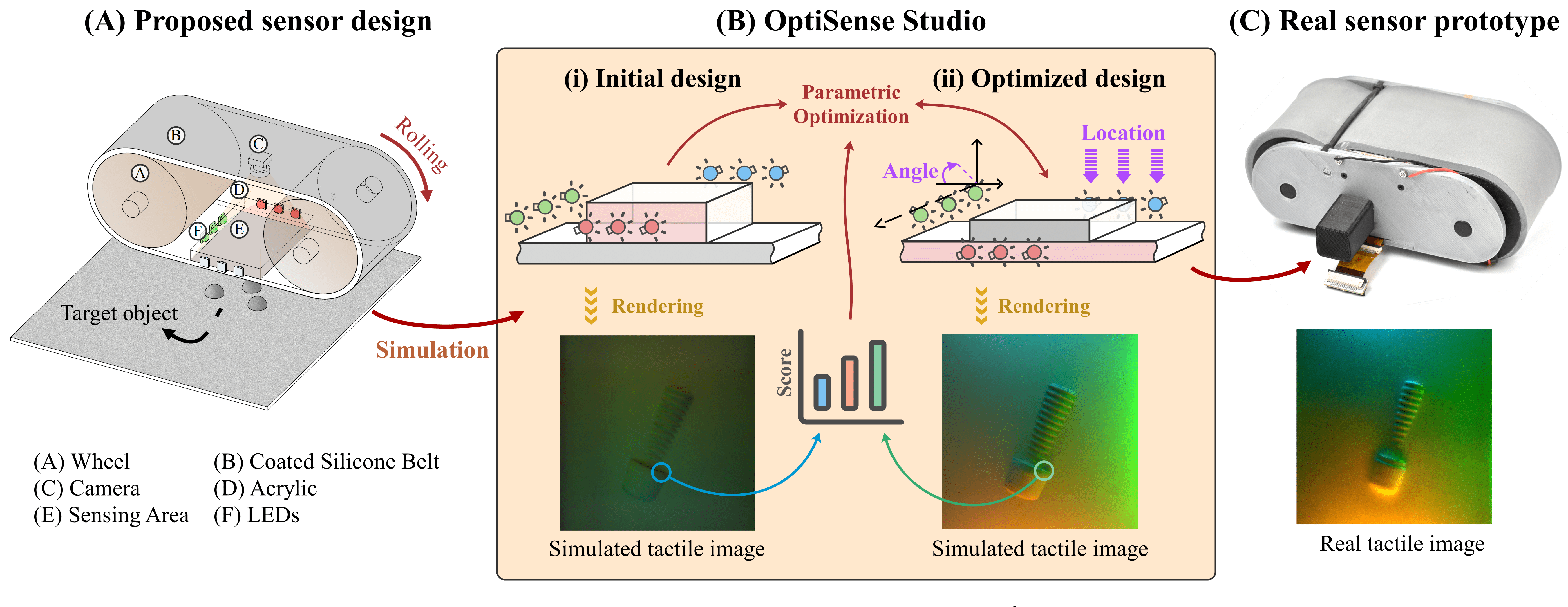}
    \caption{\textbf{Framework for modularizing and parameterizing camera-based sensors}: 
    In (\textbf{A}), the user designs an initial CAD that contains the locations and shapes of the sensor components. \textbf{(B)} This CAD design is converted to a complete sensor design in OptiSense Studio. We automatically parameterize the optical design and optimize it using novel objective functions in our simulation-driven framework. \textbf{(C)} The final design can then be directly manufactured and tested in the real world.
    }
    \label{fig:teaser}
\end{figure*}

Touch sensing enables dexterous manipulation and perception of object properties, such as hardness and textures. Vision-based tactile sensors (VBTS), specifically GelSight~\citep{yuan2017gelsight}, offer high-resolution tactile signals that has significantly improved robotic grasping~\citep{calandra2018more}, dexterous manipulation~\citep{qi2023general}, dense object shape perception~\citep{suresh2022shapemap}, and incipient slip detection~\citep{slipdetection}.

Vision-based tactile sensors use an embedded camera to capture optical cues from the deformation of the sensor's soft surface, allowing for the measurement of geometry and force at the contact surface. Among them, the GelSight sensors employ photometric stereo algorithm to measure the surface normal field, which is then used to reconstruct the 3D shape of the contact surface. This process requires a meticulously designed illumination system that evenly illuminates the surface with lights of multiple colors from different angles and controls the surface's reflectance properties by coating it with specific materials. The design space of GelSight sensors encompasses a wide range of features, including the illumination system~\citep{li2014localization,taylor2021gelslim3,finrayliu2022gelsight}, curvature~\citep{brandenroundsensor,gelfinger,tippur2023gelsight360}, mirrors~\citep{gelwedge}, and grippers with compliance~\citep{babyfinray}. In practice, the design of the optical system predominantly influences the quality of the sensor’s measurements.

A major challenge for VBTS sensor developers is customizability. Integrating a VBTS sensor into a new robot often necessitates a complete redesign of the delicate optical system. This design process is highly heuristic, with the relationship between the arrangement of individual optical components and the final sensor performance being both nonlinear and implicit. 
This creates a high barrier for roboticists who may want to customize these sensors for their target application.  
Furthermore, VBTSs introduce new design challenges because there is no standardized tactile optical component, target sensor shape, or design tool in the community. This means that to experiment with or develop such sensors, users need to design, fabricate, and implement all aspects of the tactile sensor system, which can take months or years.
We identify that there is a need to distill the human design expertise into a sensor design methodology. This will simplify the tedious trial-and-error process for sensor design and enable the exploration of new designs with rapid prototyping. 

To address the challenges, we propose a systematic design approach for GelSight family of sensors that defines the sensor design process using several parameterized optical modules. We also establish several quantitative metrics to evaluate sensor performance. We then use a physically-based simulator to assist the design, where we can optimize the design parameters based on the evaluation metrics.
Based on the methodology, we develop OptiSense Studio to enable nonprofessional users to rapidly prototype GelSight-like sensors.
OptiSense Studio guides novice users through a well-defined pipeline to form an optimal sensor design. In the toolbox, users can define the optical components in the sensor and simulate the sensor to get an intuitive understanding of the sensor output. The design parameters can be modified in an interactive interface or obtained through an automated optimization process. Even without professional optical knowledge, a novice user can easily model optical components following the interface of the OptiSense Studio and the predefined material library. The software can help users to shorten the sensor design process from months to several hours.

The key contributions of the work are:
\begin{itemize}
\item We provide a formal design procedure to generate a full GelSight-like sensor design. The framework gives a parameterized design to optimize all the sensor components and 
allows the sim-to-real transfer of optimized designs.
\item We propose four objective functions to quantify the performance of a VBTS sensor design. 
\item We created an optical component library based on the commonly used optical components in the community.
\item We implement our design framework into an interactive design tool using open-source libraries. 
\item We demonstrate the forward and optimization-driven inverse parameter selection for multiple VBTS sensor designs. 
\end{itemize}

Using our framework, we create virtual models of four different GelSight-like tactile sensors with varying challenging design features, such as light piping, curved sensing surfaces, and mirrors. Through our framework, we optimize the shape (mirror and refractive resin surface), optical material (specularity of coating material), and illumination using the parameterized designs modeled in our framework. All optimization procedures are performed virtually and can be performed in hours.  Consequently, the sensor design process is significantly speeded up. 

The modular design pipeline using virtual tools will lower the entry barrier for roboticists and help roboticists build intelligent robots. 
We believe that OptiSense Studio will greatly speed up the development of tactile sensors for new robots and boost the research and application of tactile sensing technologies for robots. 
In addition, we believe that the modularized design process, the optimization method, and the software platform can be applied to develop other optical-based sensors or tactile sensors based on other transduction methods.

\section{Related Work}
In this section, we review the designs of GelSight-like tactile sensors and virtual robotic design frameworks. We highlight the various choices in the design spaces in the most relevant sensors and highlight the connection with our design framework. Next, we discuss some work that attempts to simplify the design of vision-based tactile sensors in a limited context. 

\begin{figure*}[ht]
    \centering
    \includegraphics[width=\linewidth]{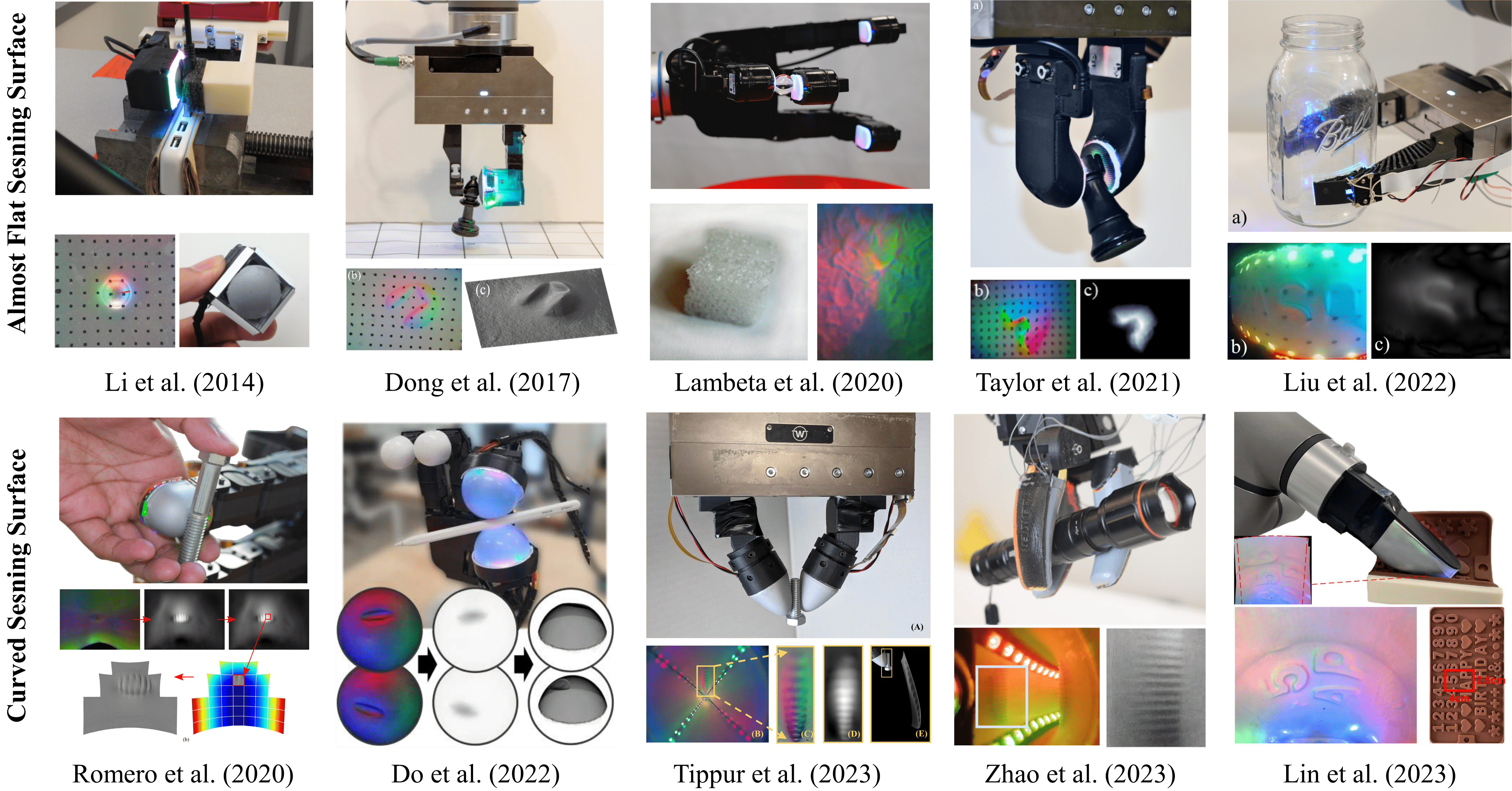}
    \caption{\textbf{Various types of GelSight sensor designs} (Pictures adapted from   \cite{li2014localization,slipdetection,lambeta2020digit,taylor2021gelslim3,finrayliu2022gelsight,brandenroundsensor,densetactv1,tippur2023gelsight360,sveltezhao2023gelsight, gelfinger}): Researchers have designed GelSight sensors with varied optical systems to fit robot fingers with either flat or curved surfaces. The performance of the sensors also varies and is affected by a number of parameters of the optical system. Our work aims to bridge the knowledge gap between expert sensor designers and novice users, thereby simplifying and expediting the sensor design process. 
    }
    \label{fig:sensors}
\end{figure*}

\subsection{Vision-based tactile sensors}
We focus on the GelSight family of sensors for our design framework, whose application to robotics was first introduced in \cite{li2014localization}. \cite{yuan2017gelsight} provides a detailed description of the manufacturing and working principle of GelSight. Since then, there has been an explosion of vision-based tactile sensor designs for various use cases~\cite{patel2021digger,gelfinger,brandenroundsensor,lambeta2020digit,tippur2023gelsight360,do2023densetact} and the co-design of grippers with integrated tactile sensing~\cite{finrayliu2022gelsight,babyfinray,sandraExoGripper,sandraEndoFlex}. 
\Cref{fig:sensors} shows an array of vision-based sensors with various design changes. 

In \cite{li2014localization}, the authors created the sensor in a rectangular shape with four sides illuminated by four different colors: red (R), green (G), blue (B), and white (W). They added light-guiding plates to pipe the light such that it enters parallel to the sensing surface. The tested sensing surface shapes were flat and dome, coated with a semi-specular material. Although the authors described the design well, it is unclear which components to use or how to approach a new sensor design after acquiring components. Our work incorporates design modules extracted from these papers and component manufacturers to allow for a quick design process. 

In \cite{slipdetection}, the authors designed a hexagonal plastic tray and mounted collimating lens LEDs at $71^\circ$ to the sensing surface. The light sources of three colors: red (R), green (G), and blue (B) were placed alternatively along the hexagonal sides. They used diffuse optical material for sensing surface coating. With the new design, it was clear that the illumination system could significantly improve tactile perception. However, it did not answer the question: What is the design space of the sensors? Is this the optimal design, given the design space? Our work addresses this gap by providing specific design spaces for GelSight sensors and optimizing the design using automated pipelines.

In \cite{brandenroundsensor}, \cite{tippur2023gelsight360}, and \cite{do2023densetact}, authors created sensors with curved tactile sensing surfaces for dexterous manipulation. Exploring the design space of shapes by manufacturing sensor prototypes is prohibitively expensive and time-consuming. Our design framework allows users to perform sensing surface shape exploration using simulation-driven and objective-driven pipelines. 

\subsection{Virtual robotic design and Sim2Real}
In this section, we cover works that leverage simulation for sensor design and other Sim2Real works that focus on robotic manipulation.   

\cite{taylor2021gelslim3} is the most similar to our work. In this paper, the authors redesigned the optical system of GelSlim~\citep{gelslim1} to uniformly illuminate the sensing surface and recover the surface geometry. They used raytracing simulation software to design a shaping lens and LED light position. Their simulation-driven approach was useful for coming up with a non-trivial lens shape. Although the authors used simulation, the output modality (radiant flux) was different from the tactile image (camera image). The raytracing software used by the authors is more focused on professional optical designers and requires detailed models of optical components, which can be overwhelming for roboticists. Moreover, their design tool did not provide any guidance on how to generate tactile sensors or provide any objectives for automatic parameter selection. 
Therefore, it is unclear whether their approach can be extended to an end-to-end approach for tactile sensor design. 

In \cite{zhang2024pfs}, the authors analyze the common design pipelines of vision-based sensors and propose a dictionary-based process flow design approach. Their approach is useful for mixing and matching various workflows for sensor manufacturing. Although useful, their work does not provide any feedback on the validity or sensing ability of the design. Our work is the first to provide simulation-driven interactive feedback on the validity and perception capabilities of sensor design. Moreover, their work does not consider any optimization-based parameter selection techniques.

In \cite{graspStabilitySim2Real}, the authors use efficient tactile simulation to train a grasp stability model completely in simulation and show zero-shot transfer to real robots. This approach depends on Taxim~\citep{si2022taxim} which requires data from the specific real-sensor prototype for simulation. Since our focus is on creating new sensor designs virtually, this simulation approach is not applicable to our problem.

\section{An overview of the design framework}
\begin{figure*}
    \centering
    \includegraphics[width=\textwidth]{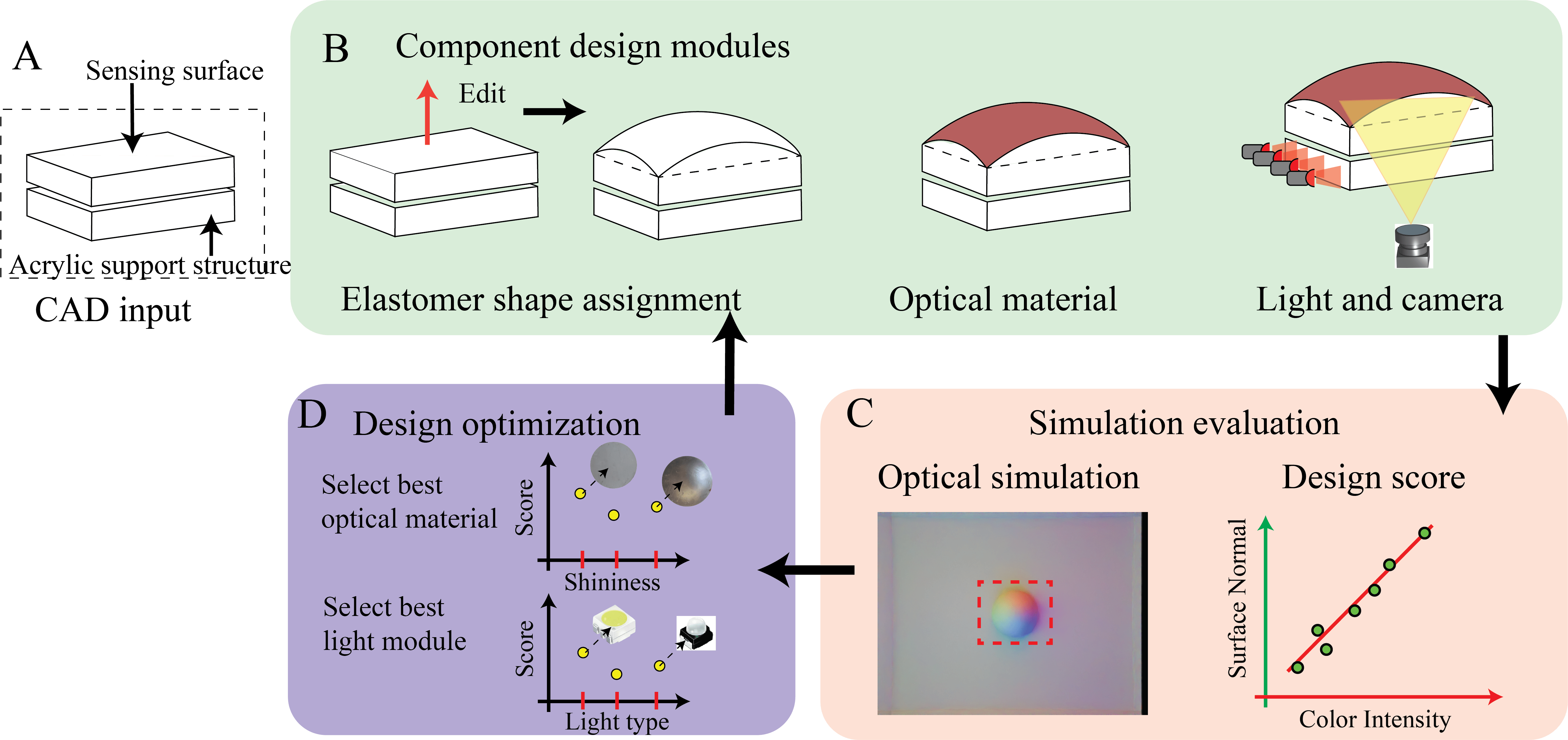}
    \caption{\textbf{Sensor design framework}:
    Given the user shape input in A, we model the sensor design with multiple modules in simulation as shown in B. We then evaluate the sensor performance based on the simulated indentation test in C. This is then coupled with optimization methods to choose the optimal light module and optical coating material for the sensor design.
    }
    \label{fig:framework}
\end{figure*}

We introduce a formal design process for GelSight-like tactile sensors. The method is based on modularizing the design of the sensor's optical system into several key optical modules, parameterizing them, and designing a set of evaluation metrics for the sensor design. 
The process starts with a user inputting the initial sensor design and optical component for parameterization, one sensor design can be represented by the parameters of the optical components. 
We perform the modeling and optimization process using a simulation platform based on physics-based rendering (PBR), which models the optical characteristics of individual components. We then perform an optimization process in simulation based on the evaluation metrics to find the best design parameters and finally transfer the designs to a real-world prototype. The pipeline is demonstrated in \Cref{fig:framework}. We further discuss the design parameterization in \Cref{sec:designparam} for geometric shape, optical material, and light source; the evaluation metrics procedure in \Cref{sec:objectivefunc}; and the implementation of the system in \Cref{sec:OptiSense}.

\subsection{GelSight sensor modularization} \label{sec:modular}
We decompose sensor modeling into five parts: \textit{Soft elastomer}, \textit{Support structure}, \textit{Opaque coating}, \textit{Light}, and \textit{Camera}. For each part, we introduce a design module to create and optimize the corresponding part. Each part can be either modeled from scratch or initialized from our component library. We distill common optical components based on vision-based sensor literature into a component library (see \Cref{app:componentsLib}). This enables novice users to model sensors without any experience with VBTS sensors. 

\begin{figure}
    \centering
    \includegraphics[width=\columnwidth]{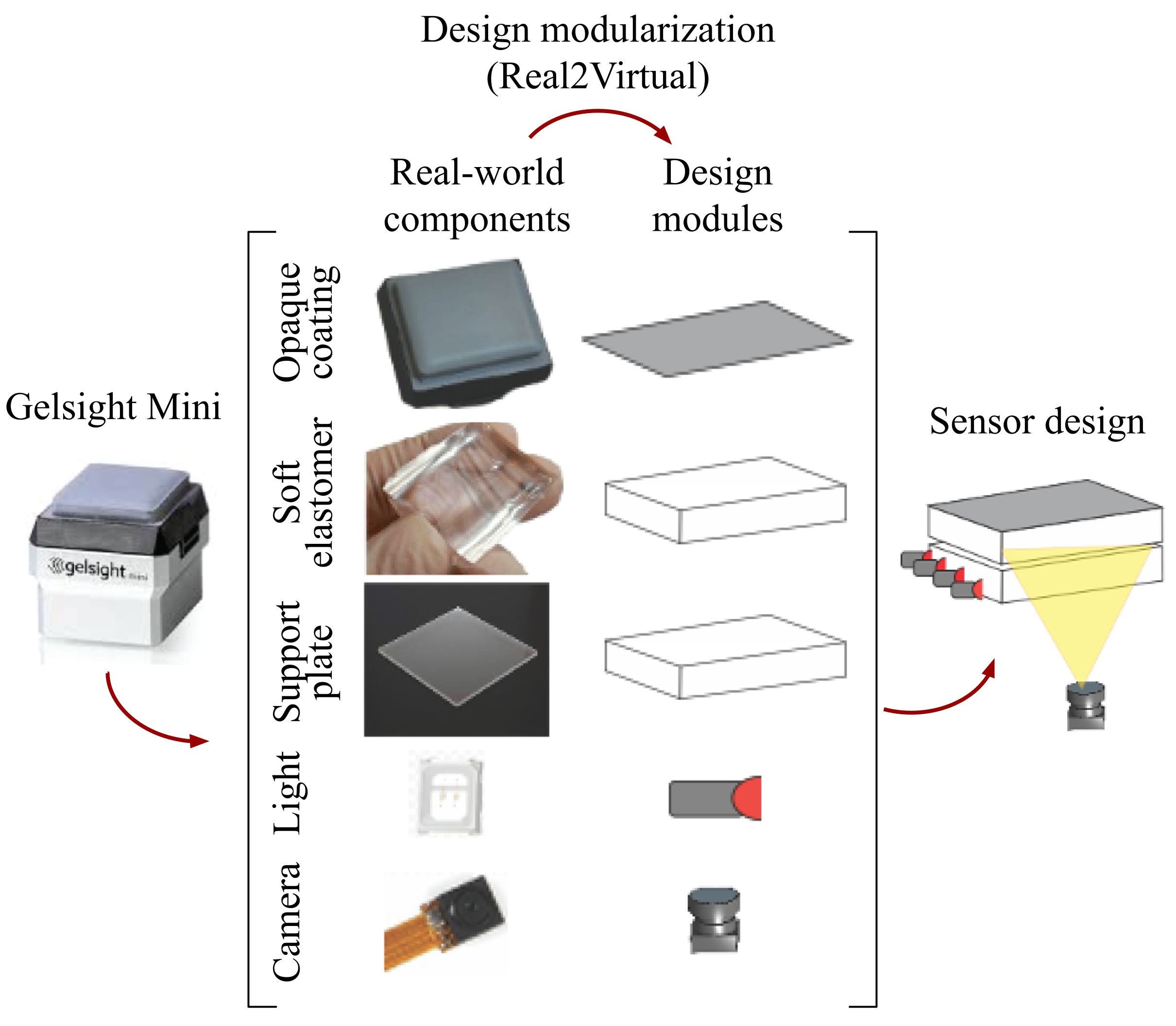}
    \caption{\textbf{Sensor modularization}: This figure illustrates how a tactile sensor can be modularized into our proposed modules. These modules can then be used to create a digital design for further optimization.}
    \label{fig:modularize}
\end{figure}

\Cref{fig:modularize} shows an illustration of how a GelSight Mini tactile sensor could be decomposed into real-world components. These components are analogous to the parts (modules) in our design framework. We use these modules to create a digital design that can be subsequently optimized. 

The key elements of each module are the choice of reference surface for the shape description, the selection of optical materials, and the choice of discrete elements from the component library. 
For a short tutorial on the modules, see \Cref{app:guiTutorial}.

\subsection{Design optimization}
\label{sec:optimFramework}

Given the design parameterization, the user can leverage the forward and inverse design process for parameter selection (parameters include light types, coating material, and sensor surface geometry).
For the forward design process, the user can select any parameter, manually change that parameter in OptiSense Studio, and evaluate the design. We found that the light location setting is one parameter that can benefit from this approach. We show experiments in \Cref{sec:rollerDesignExp} to optimize light location using this approach. 

In the inverse design process, the user can perform an automated search over the parameterized space. We provide two methods, discrete search and gradient-free optimization, to select the best design. Specifically, we use grid search over the parameter space for discrete optimization and CMA-ES~\citep{hansen2016cma} for optimizing over continuous parameter space. The user can choose the type of optimization method based on the time available for design. CMA-ES takes more time than grid search, but CMA-ES can return a design with a better objective function score. We show experiments that use both of these techniques for optimizing designs. 

We discuss the specific parameterization of various optical components in \Cref{sec:designparam}.

\section{Design parameterization} \label{sec:designparam}

In this section, we describe the parameterization of the key sensor components. This will be the basis for sensor design modification. The choice of parameterization was made on the basis of the authors' expertise in VBTS tactile sensor design. 

\begin{figure}
    \centering
    \includegraphics[width=\columnwidth]{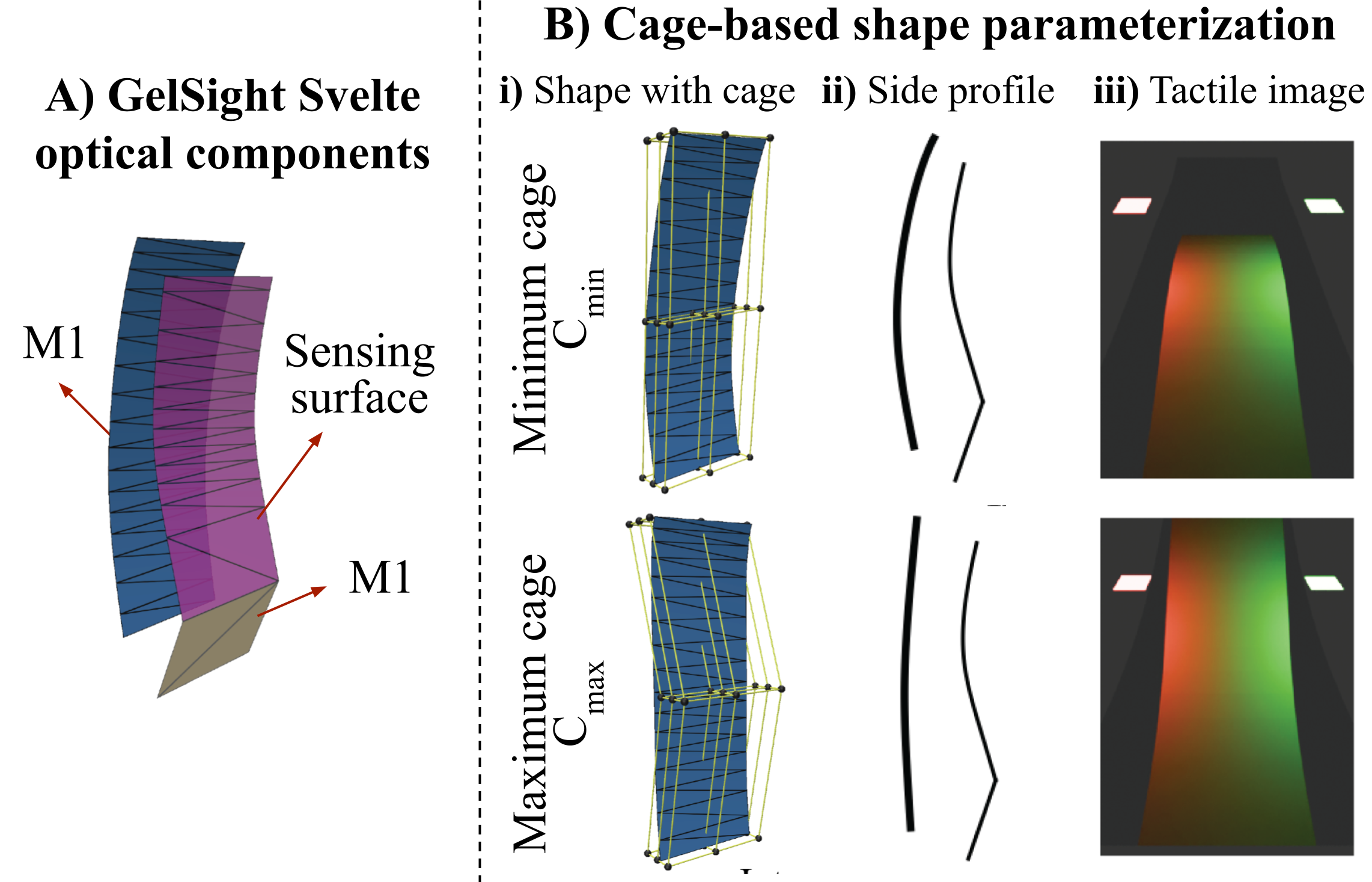}
    \caption{\textbf{Shape parameterization}: We use a cage (the bounding box of the mesh surface) to parameterize the component shape, shown in the left column as \textit{M1} representing mirrors in GelSight Svelte. The cage is set as the geometrical parent for the target surface mesh (child). Changing the shape of the cage consequently changes the mesh.
    The right column shows the user input of optimization boundary $\mathcal{C}_{\text{min}}$ and $\mathcal{C}_{\text{max}}$. We show the deformed surface, 2D profile, and the corresponding tactile image to qualitatively represent the change in tactile signal by changing \textit{M1} mirror element in the sensor. Therefore, shape optimization of \textit{M1} is critical to obtaining the best sensing performance.}
    \label{fig:shape_optim}
\end{figure}

\textbf{Geometric shape of the components.}
All the components in the sensor are geometrically represented by triangle meshes. 
Therefore, the shape representation of one part can be of very high dimensionality, such as in the order of magnitude of $10^4$. 
To find a low-dimensional parameterization of the shape, we apply a cage-based representation of the meshes inspired by \cite{xu2021end}.
An example of the representation is shown in \Cref{fig:shape_optim}: given a complicated shape in mesh, we automatically generate a cuboidal cage with 27 cage vertices such that the cage completely encloses the component in the form of a bounding box. We then use the cage vertices to represent the entire mesh. We can increase the resolution of the cage interactively if more fine-grained control is desired. The cage-based representation can be applied to any surface mesh irrespective of how it was generated (B-Rep representation~\cite{brep_stroud2006boundary}) and bounds the dimensionality of the optimization problem for automatically choosing the component shapes. 

For shape optimization, we require users to specify the boundary of the cage deformation specified as $\mathbf{C}_{\text{min}}$ and $\mathbf{C}_{\text{max}}$.
Then the current shape is given as 
$$\mathbf{C}_{\text{current}} = (1 - \mathbf{A}) \cdot  \mathbf{C}_{\text{min}} + \mathbf{A} \cdot \mathbf{C}_{\text{max}}$$
where $\mathbf{C}_{\rm{min}}, \mathbf{C}_{\rm{max}}, \mathbf{A} \in \mathbb{R}^{N\times3}$ and $\mathbf{A}_{i,j} \in [0,1]$.
In our implementation, the shape representation uses 27 cage vertices, $N=27$.

\textbf{Optical material.} 
In a VBTS, the user needs to assign material to optical components, like, coating surface, elastomer, and supporting structure. 
Generally, optical material properties are characterized by a bidirectional scattering distribution function (BSDF), a 4-D function parameterized by the incoming and distribution of the outgoing rays. However, specifying or measuring a full BSDF model is extremely challenging \citep*{adaptivebsdfparam}. Therefore, we leverage physically motivated \textit{analytic} BSDF models proposed in the computer graphics community \citep*{Mitsuba3-BSDFs} for modeling the vision-based tactile sensors. The relevant optical material models are \textit{RoughDielectric}, \textit{RoughConductor}, and \textit{Diffuse}.
We performed optical experiments to calibrate these models for various surface materials (sensing surface coating and transparent surfaces) in real sensors. The detailed experimental setup and calibration results for coating material and light sources are available in the supplementary document. For a detailed description of the optical materials, see \Cref{app:componentsLib}. 

\textit{RoughDielectric} can be used for all refractive and transparent surfaces in the sensor, like elastomer and clear support structure. This model has two relevant properties: the refractive index ($\eta$) and the roughness ($\rho$). We obtain calibrated values for the elastomer (PDMS) and the resin (epoxy) in our component library.
We use the \textit{RoughConductor} model for representing the sensing surface opaque coating material. This model has seven relevant properties: RGB reflectance (3D), refractive index ($\eta$) (3D), and specularity $\rho$ (1D). We can synthesize all the relevant coatings used in the GelSight family by varying the 1D specularity property of this model. Therefore, for material optimization, we use \textit{RoughConductor} optical model and vary the specularity ($\rho$) value $[0,1]$ to obtain the best coating material for the specific sensor design. 

\textbf{Light source.} We allow changes to the location, orientation, and type of the light sources.
For sensor design, the user can select various types of light available on the market. These light types mostly differ in their outgoing ray distribution and color. The outgoing ray distribution is characterized by an IES profile and is available from the light manufacturer. 
For light sources, we use the modified \textit{PointLight} and \textit{AreaLight} models from Mitsuba~\citep*{jakob2010mitsuba}. 
Specifically, for lights that have a spherical lens, we leverage the IES light profile provided by the manufacturer and add it to \textit{PointLight} to scale the intensity value along a specific outgoing direction. For LEDs with a flat lens, we use the dimensions provided by the manufacturer and scale the \textit{AreaLight} accordingly to provide an approximate model. These light models are based on calibration experiments of real LEDs, specifically topled OSRAM LEDs and Chanzon 5730 LEDs. Please refer to the supplementary for detailed calibration setups and sim2real comparison of the light models. For a detailed description of the light sources, see \Cref{app:componentsLib}. 

For light design, we modify the \textit{location} and \textit{orientation} of the light group
using a forward design approach. GelSight-like VBTS sensors have two or three light panels that contain lights of the same type. We call these similar light collections a \textit{Light Group}.
We group lights based on their color. For light optimization, we vary the \textbf{light type} for all lights in the design.  Specifically, we consider 3 types of variations: a) single light type (area vs. point) for all the lights in the sensor model (number of parameters=1); b) single light type for all the lights in a group (number of parameters=number of light groups; generally number of light groups=3); c) change the type of each light individually (number of parameters=number of lights in the sensor; can vary from 15 to 30). The last variation has all the possible configurations, including a and b.

\textbf{Camera.} We choose the perspective camera model in all designs. We allow three parameters, height, width, and field-of-view (FoV) of the camera, to be varying. For a detailed description of the camera see \Cref{app:componentsLib}. For optimization, we iterate through the list of available cameras in our library and set the corresponding parameters to evaluate the sensor design. 

To simulate the camera's artifacts of saturation, we post-process the rendered images. Initially, we export the rendered image in OpenEXR format, akin to a RAW format in real cameras, capturing the light intensity for each pixel independent of the camera's exposure settings. Subsequently, we convert the image to PNG format based on the desired exposure value. To approximate artifacts observed when capturing bright areas—such as regions near LEDs—we mask the overexposed areas and apply a Gaussian blur to them. We then propagate the saturated pixel values across all three RGB channels according to the mask pixel intensities. This process results in whitish spots on light sources and saturation in areas adjacent to the lights.


\section{Objective functions for sensor design} \label{sec:objectivefunc}
In this section, we introduce four evaluation metrics for sensor performance. These metrics will be used as objective functions during the design optimization procedure. The metrics will provide quantified guidance on ``what a good sensor should perform like'' instead of the human experts' assessment. Note that sensor designers need to highlight performance from different perspectives, such as precise geometry measurement or smaller distortion in measurement, for different design purposes. Therefore, we propose multiple objective functions for different optimization goals. So, in the design phase, users can choose one or multiple objective functions based on specific design goals. 

\subsection{RGB2Normal mapping objective function}\label{sec:RBG2NormalFunc}
GelSight sensors family works based on the example-based photometric stereo technique to recover high-quality surface normals, and then integrates the surface normals to obtain the 3D surface topography \citep{johnson2009retrographic}. The key calibration step is to create a mapping between the RGB color and surface normal, and the accuracy of the color-to-normal mapping is the key to the accuracy of 3D reconstruction.
Our first metric quantifies the sensor's shape measurement accuracy by evaluating how well it can recover the surface normals based on the color. In most cases, we can assume that the local color-to-normal mapping is independent of the contact shape, but could be influenced by the location of the measurement.

\begin{figure}
    \centering
    \includegraphics[width=\columnwidth]{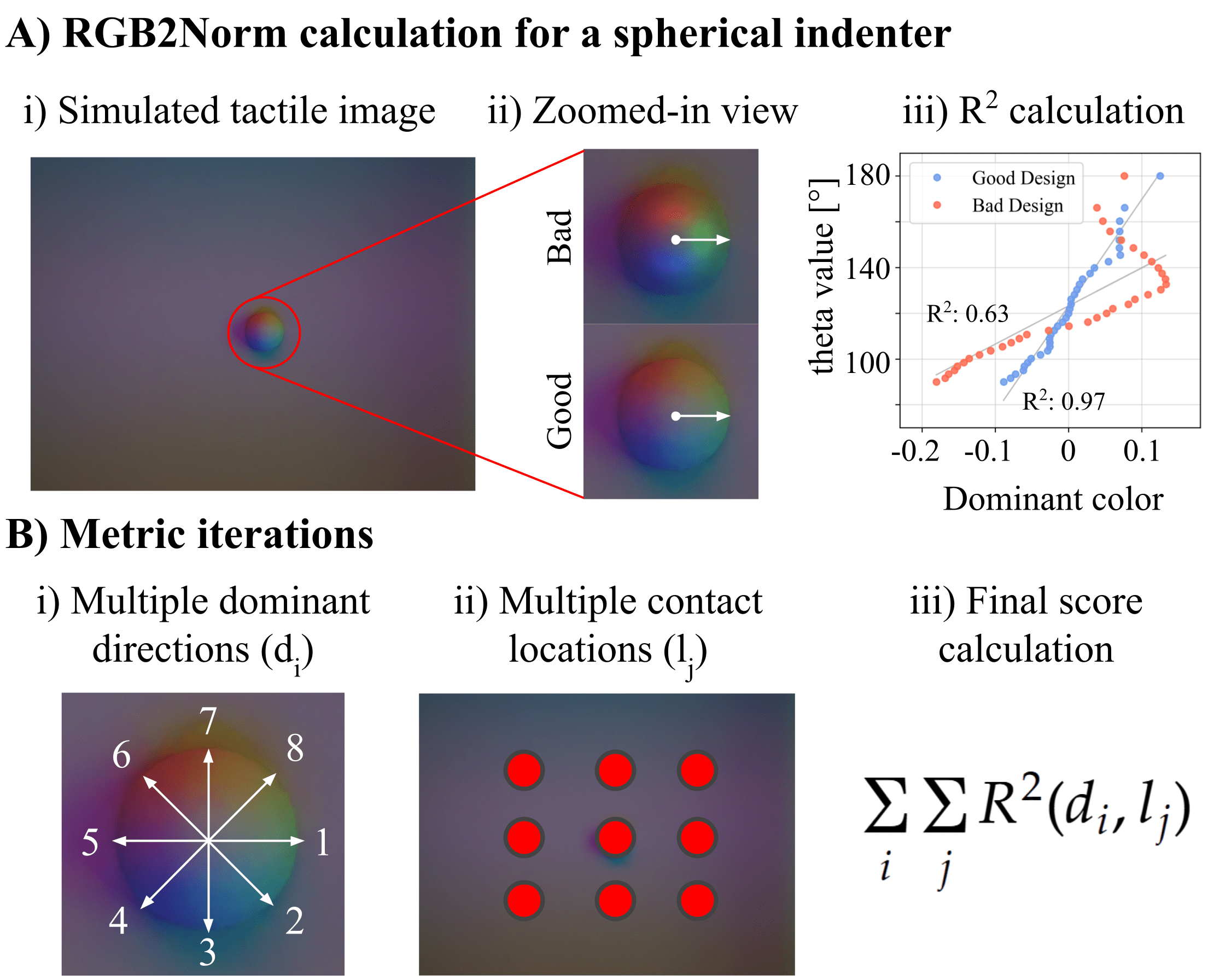}
    \caption{\textbf{Evaluation criteria method}: In \textbf{A}, we show the linearity fit calculation for a single indenter location. We use the $\theta$ value of surface normals and dominant color for calculating the linearity score. We average the score across multiple dominant directions (\textbf{B}). To account for spatial variation in our evaluation criteria, we average the value across multiple contact locations one by one. The final calculation is given in \textbf{Biii}.}
    \label{fig:evalCriteria}
\end{figure}

We choose to test the mapping between the RGB value and surface normal using spherical indenters at different locations and average the mapping quality. As hemispherical contact area provides good coverage of all surface normals, spherical indenters are commonly used for calibrating GelSight sensors. \Cref{fig:evalCriteria} shows a pictorial representation of the design evaluation process, which we sample along different directions in the indentation area to measure the correlation between the change of the surface normal values and the RGB values. 
As noted in \citep[Section 5.1]{yuan2017gelsight}, to obtain good surface reconstruction results, the two should have a strong correlation and ideally be linear. 

The evaluation procedure is as follows: 
\begin{enumerate}
    \item Generate two tactile images: without any indentation and with spherical indenters.
    \item Select a line segment in the indented region passing through the center of the sphere.
    \item Obtain the dominant color by performing dimensionality reduction using PCA~\citep{abdi2010principal} on the 3D color values along the line.
    \item Plot the dominant color versus surface normal $\theta$ (polar coordinate) value and calculate the linearity score ($R^2$-score). To calculate the $R^2$-score, we used \textit{linregress} from the scipy library~\citep{2020SciPy-NMeth}.
    \item Repeat the process for other directions ($d_i$). We chose eight directions (cardinal and ordinal directions) to capture variation.
\end{enumerate}

To capture the effect of spatial variation, we repeat this process for multiple locations ($l_j$) and average the score, as shown in \Cref{fig:evalCriteria}B. In our implementation, the default number of indenter locations is nine and the location of indenters is chosen to cover the sensing surface uniformly. The radius of the default spherical indenter is \qty{1.5}{\mm}.
The final equation for the evaluation criteria is shown in \Cref{fig:evalCriteria}D. 

This objective function is fast to calculate as it requires minimal extra computation (normal rendering and linearity fitting) and serves as a good proxy for 3D reconstruction quality of the tactile sensor. 

\subsection{NormDiff objective function}
We introduce the \textit{NormDiff} objective function as an alternative function to evaluate the sensor's capability of measuring 3D shapes. The motivation of the function design is the same as the RBG2Normal function design introduced in \Cref{sec:RBG2NormalFunc}, but here, we do not use the constraint that the RGB vector is expected to be linear to the surface normal value. Instead, we expect that the RGB value corresponding to a specific surface normal vector should be very distinct from the ones of other surface normal values. The level of ``distinctiveness'' is denoted by the measurement uncertainty, and we calculate it based on the prominent camera noise models~\cite{pagnutti2017laying}. In this model, the RGB noise is proportional to the sensor response or the RGB value at each pixel.

\begin{figure}
    \centering
    \includegraphics[width=\columnwidth]{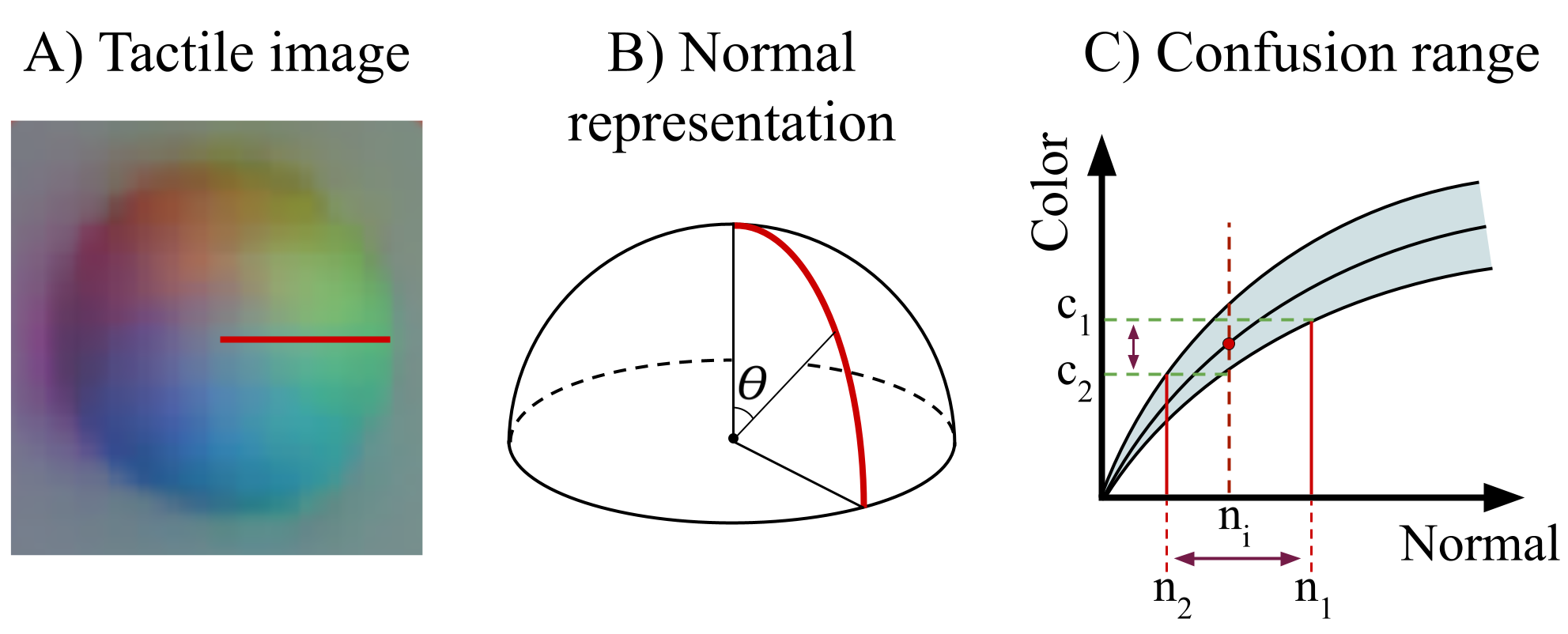}
    \caption{\textbf{NormDiff objective function:} (A) shows the tactile image with an indentation; (B-C) shows the canonical example of color-normal plot. For a chosen normal, $n_i$, the color noise between $[c_2, c_1]$ leads to a confusion range in normal to be $[n1-n2]$.}
    \label{fig:NormDiff}
\end{figure}

For an intuitive explanation with a canonical illustration, refer to \Cref{fig:NormDiff}. In the canonical example, the color and normal are both represented as 1D quantities. The color-normal curve along with the standard deviation is shown in \Cref{fig:NormDiff}B. At normal value $n_i$ the color values vary between $c_1$ and $c_2$. For color values $c_1$ and $c_2$, the range of normal values is $[n_i, n_1]$ and $[n_2, n_i]$ respectively. Therefore, at the normal value $n_i$, the confusion in estimating normal is $(n_1 - n_2)$ and \textit{NormDiff} objective function value of an indenter is the negative of the confusion, $-(n_1 - n_2)$. 

In the original implementation, we used the lookup table (LUT) to create a mapping between the 3D RGB color values $(r_n,g_n,b_n)$ and the 2D $(\theta, \phi)$ surface normal coordinates, as this is common practice in GelSight sensors~\cite{yuan2017gelsight}. To calculate the confusion value per indenter, we follow these steps.
\begin{enumerate}
    \item We identify indented pixels by using the difference between surface normals before and after indentation. For each pixel, we record the $(r_n,g_n,b_n,\theta_n,\phi_n)$ tuple to create a dataset.
    \item To assess the recovery quality for a surface normal, $(\theta_i, \phi_i)$, we find the nearest tuple in our dataset, $\mathcal{R}_1 = \{p_j\}, j=1(1)U$, among the indented pixels. 
    \item For each data-point, $p_j$, we have the corresponding tuple, $(r_j,g_j,b_j,\theta_j,\phi_j)$. In the color space, we obtain the range of color values by adding noise to the RGB value, $(r_j,g_j,b_j)$. In our implementation, we added noise that was \qty{30}{\percent} of the RGB value based on heuristic measurement of many consumer cameras. Note that a different choice of the value can lead to very similar optima.
    We find all the data points, $\mathcal{N}_1 = \{q_k\}, k=1(1)N$, in the color space within this RGB range. 
    \item Each data-point, $q_k$, has its corresponding normal value, $(\theta_k, \phi_k)$. We calculate the maximum and minimum values of the $\theta$ and $\phi$ values between all the points in $\mathcal{N}_1$. The confusion for each data-point, $p_j$, is then the weighted sum of the range $\theta$ and the range $\phi$. 
    \item We take an average across all the data-point ($p_j$) to obtain the confusion to recover surface normal, $(\theta_i, \phi_i)$.  
    \item We repeat Steps 2-5 for other $(\theta_i, \phi_i)$ pairs. 
    \item The final value of the objective function is negative of the average of the confusion from the previous step.
\end{enumerate}

\subsection{As-orthographic-as-possible (AOAP) objective function} 
For complicated GelSight sensor designs, the camera rays are not perpendicular to the sensing surface, causing distortion of the geometry at the contact surface. This geometrical distortion is not desired for sensing and can be decreased by changing the sensor shape. 
We introduce an objective function to measure this distortion systematically for optimizing the sensor through shape parameterization explained in \Cref{sec:designparam}. The key idea of this objective function is that the angle of incident rays when they reach the sensing surface should be as close to zero as possible. If the sensing surface is a plane, then this condition would make the rays as if they were shot from an orthographic camera. Therefore, we name this objective function \textit{as-orthographic-as-possible} (AOAP). We also add a regularizer term to this objective function to encourage sensing surface coverage to avoid mode collapse. 
The calculation procedure is as follows: 
\begin{enumerate}
    \item Shoot rays from the cameras and perform ideal refraction and reflection on surfaces until the rays hit the sensing surface or escape the sensor. Record the hit position, hit triangle face index and incidence angle of all the rays. 
    \item Find unique sensing surface mesh faces, $U$, hit by all the camera rays. The total number of mesh faces in the sensing surface is given by $T$.
\end{enumerate}
The final objective function is as given below:
$$\mathcal{O}_3 = \frac{1}{N}\sum_i\mathbf{n}_i \cdot \mathbf{\omega}_i + k_1 \frac{U}{T}$$
where $k_1=0.01$ and N is equal to the number of pixels in our experiments.

\subsection{2D-to-3D projection warping (2to3PW) objective function}
VBTS tactile sensors capture a 3D sensing surface in a 2D camera image. These sensors generally consist of refractive components like elastomers and reflective components like mirrors. The 2D image field is warped because of the presence of these optical components. We introduce an objective function to measure this warping. The main intuition is that each pixel is a planar square with the same area. Each pixel's image on the sensing surface should remain close to a square of the same area. This objective function was developed to capture the warping observed in highly curved tactile like in thumb designs in \cite{andrussow2023minsight}. 
The calculation procedure is as follows: 
\begin{enumerate}
    \item Shoot rays from the cameras and perform ideal refraction and reflection on surfaces until the rays hit the sensing surface or escape the sensor. Record the hit position of all the rays. 
    \item Divide each pixel into two triangles and calculate each triangle area, $\Delta_i$, in closed form.
    \item Calculate the mean ($\mu$) and the standard deviation of the triangles ($\delta$). The score value is the ratio of mean to standard deviation.  
\end{enumerate}

Note 2to3PW objective function is different from the previous objective function (AOAP), as it tries to reason about the global warping as compared to local distortion introduced by light rays.

\section{OptiSense studio: design toolbox for vision-based tactile sensors}\label{sec:OptiSense}
We implement our modular design framework in a simulation-driven design toolbox. We give a description of the design interface in \Cref{sec:interface}, give a brief description of the simulation technique used in \Cref{sec:sim}, and describe a new component library of VBTS components in \Cref{sec:componentLib}. \Cref{fig:framework} shows the whole framework. 
For a detailed description of the modules, see \Cref{sec:pipeline}.

\subsection{Design interface}
\label{sec:interface}
\begin{figure}
    \centering
    \includegraphics[width=\columnwidth]{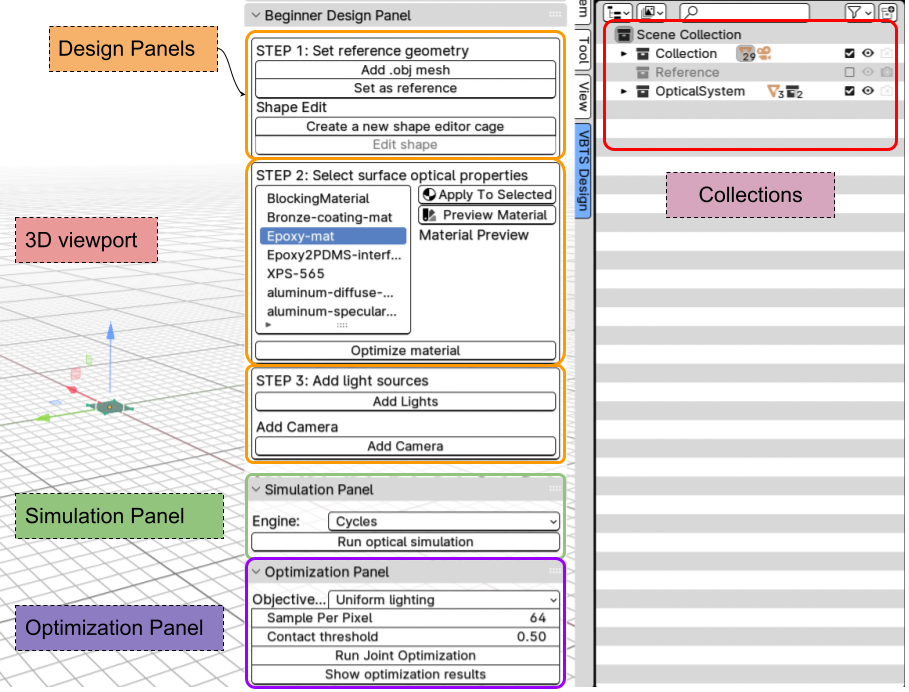}
    \caption{
    The interface of OptiSense Studio, which is built in Blender. The interface
    consists of a 3D viewport to visualize the model, various panels to perform parameterized digital design and 
    select from component collections.}
    \label{fig:gui_interface}
\end{figure}

\Cref{fig:gui_interface} shows an overview of the digital design interface. Our design environment is built on top of Blender (version 4.1.0)~\cite{blender} using its Python API for scripting. The relevant elements of the interface are a 3D viewport to visualize the design in 3D; a collection panel to group components used for modeling and optical simulation; design component modules, a simulation module, and an optimization module in the add-on panel.

\subsection{Sensor simulation}
\label{sec:sim}
We build on the simulation technique described in \cite{PBRtactilesim} to generate tactile images. The simulation technique is based on physics-based rendering (PBR) and, therefore, allows for customizable optical designs.  
We implemented the Stochastic Progressive Photon Mapping (SPPM)~\cite{hachisuka2009stochastic} rendering algorithm in Mitsuba 3~\citep{jakob2022dr} and used it to generate all tactile images. 

In path tracing, light rays are shot from a perspective camera, bounce off from various surfaces, probabilistically, and reach light sources. The surface material can have properties such as reflection, refraction, and absorption of light. PBR is the preferred technique for optical simulation because it accounts for light properties such as the principle of energy conservation and reciprocity. It can generate images after taking all the light bounces into account in the environment, also known as global illumination. It also allows to incorporate physically grounded optical components such as material, lights, and cameras. 

SPPM consists of three phases: (a) rays are shot from a perspective camera, bounce off from various surfaces, probabilistically, and deposit on a diffuse surface. The deposited points are called visible points; (b) rays are shot from light sources, bounce from specular sources, and stop at diffuse surfaces. They are called photons; (c) The contribution of each photon is added to the nearby visible point via kernel density estimation. SPPM progressively generates images with iteratively running these phases with decreasing search radius. This algorithm is specifically suited to optical simulation as we often have multiple refractive surfaces and point light sources in tactile sensors. Please refer to \cite{pharr2023physically} for a detailed description of the algorithm. 

The key benefits of the approach are that it allows the simulation of a wide variety of design spaces required for vision-based sensors and it generates physically accurate tactile images that can be used for obtaining tactile signals. 

\subsection{Component library} \label{sec:componentLib}
To allow users to quickly assemble a tactile sensor, we provide a set of commonly used optical components. 
We provide multiple cameras, physically accurate light sources, and calibrated optical materials. This library was built by extensively performing component calibration using real-world optical experiments. Users are also allowed to edit or create new optical components using the components provided in the library.
For a detailed description of the library, see \Cref{app:componentsLib}.

\section{Experiments}
In this section, we leverage our design framework to model and optimize four different types of GelSight sensors: commercial GelSight Mini~\citep{li2014localization}, GelBelt with rolling capability, omnidirectional GelSight360~\citep{tippur2023gelsight360}, and mirror-based GelSight Svelte~\citep{sveltezhao2023gelsight}. These sensors have various simulation and design challenges. GelSight Mini and GelSight360 use \textit{light piping} for uniform illumination on the sensing surface. GelSight360 and GelSight Svelte have curved sensing surfaces. In GelSight Svelte, the camera view is guided through multiple mirror surfaces to cover a curved finger-like sensing surface. We can simulate all sensors without any sensor-specific calibration. The key objectives of our experiments are that through our design framework users can easily edit previous sensor shapes (GelSight Mini), create new sensors (GelBelt), explore the design space of existing sensors (GelSight360), and optimize tactile perception by changing the shapes of optical components automatically (GelSight Svelte). We perform exploration and design optimization completely in simulation. Furthermore, we manufacture sensor prototypes for new designs of GelSight Mini, GelBelt, and GelSight Svelte to show sim2real transfer.

\begin{figure*}
    \centering
        \includegraphics[width=\textwidth]{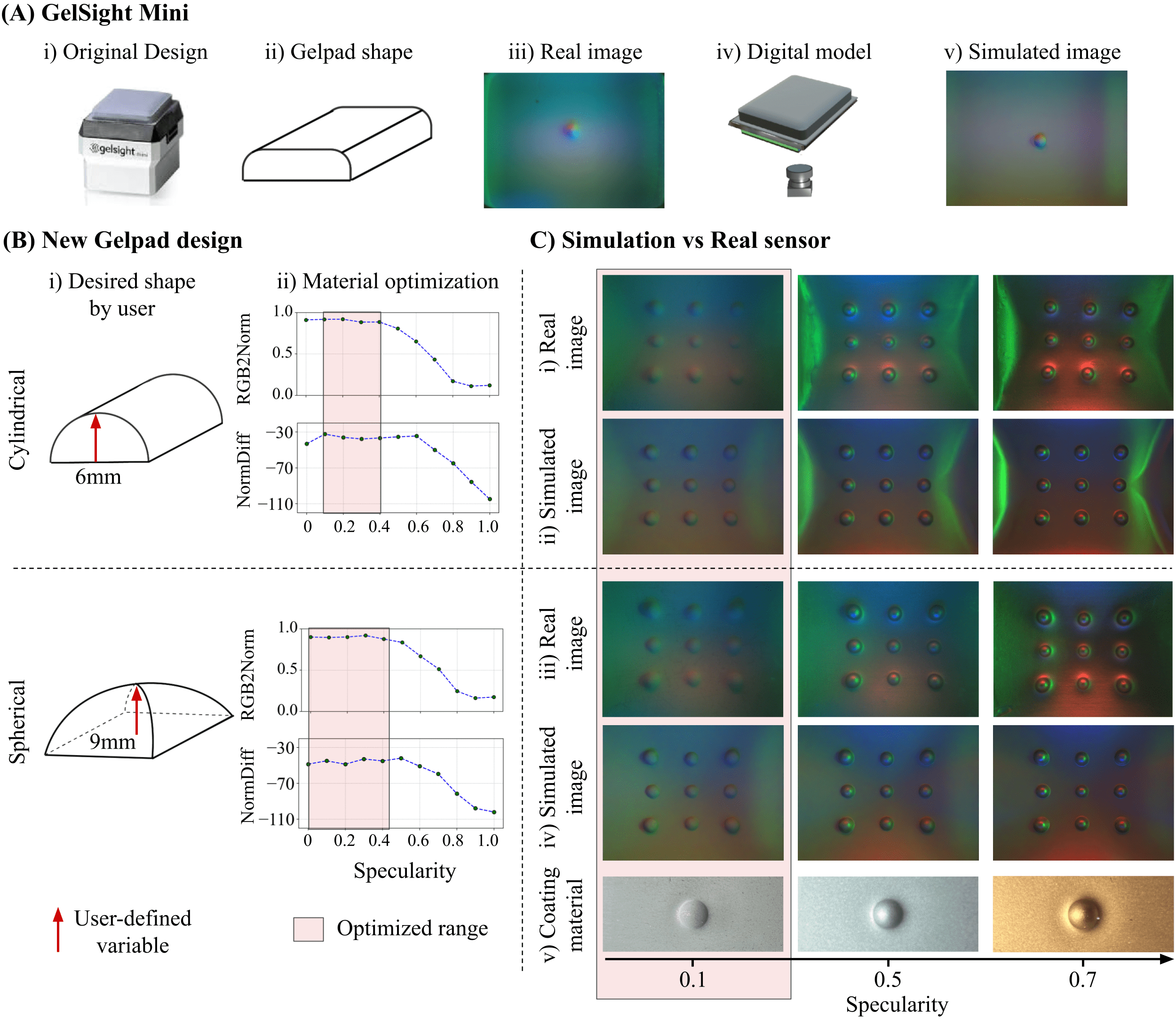}
    \caption{\textbf{Modeling and customization results of GelSight Mini}: A) shows the Gelsight Mini sensor (i) with default flat sensing surface (ii) and a real-world tactile image with a sphere indenter (iii); (iv) shows the digital design and (v) shows the simulated image for this flat design. B) We created 2 curved variants, cylindrical and spherical, by editing the initial sensing shape and showing optimization results. For each new shape, we show the gelpad shape, coating material versus evaluation score plot, optimized digital design, and simulated and experimental tactile image with 9 spherical indenters.}
    \label{fig:miniResults}
\end{figure*}

\subsection{Case study I: Curved customizations of GelSight Mini} \label{chpt:MiniExperiment}
GelSight Mini is one of the few commercially available vision-based tactile sensors and is adapted from the sensor design introduced in \cite{li2014localization}. The sensor is shown in \Cref{fig:miniResults}Ai. It allows easy integration in robotic fingers that have a flat surface. However, for general-purpose robots, there might be a need to make the sensing surface non-planar. We show the design iteration of the non-planar sensing surface
of GelSight Mini in this section. First, we assemble the initial design using our design framework. Second, we modify the sensing surface shape using our cage-based representation to the desired shape. Third, we iterate on the optical material properties of the sensing surface and light type. In our experiments, we did not find substantial improvements in sensor's measurement when optimizing the sensor's shape based on curvature. 
The simulation time to generate each GelSight Mini tactile image takes 6 seconds on an M2 MacBook Air.  

\textbf{Modeling GelSight Mini in OptiSense Studio}. 
We take the CAD provided by GelSight Inc. (original sensor vendor) and create an optical design as shown in \Cref{fig:miniResults}A(iv). Due to the availability of design optical components in our library, the optical design can be created in minutes, and a simulated tactile image can be generated. In \Cref{fig:miniResults}, we show the real and simulated sensor images after pressing the spherical indenter on the surface in parts (iii) and (v), respectively. The simulated image closely matches the real one. 

\textbf{Editing sensing surface shape}.  We modify the shape of the sensing surface by moving the control point of the cage-based representation, which is automatically generated.
To obtain the cylindrical surface, we moved the cage control points in the middle row along the z-axis by \qty{6}{\mm} (this parameter was arbitrarily chosen to make the sensor cylindrical). To obtain the spherical surface, we moved the cage control center point along the z-axis by \qty{9}{\mm} (this parameter was arbitrarily chosen to make the sensor spherical). This shows the ability of our shape editing tool to create custom sensors with curved shapes for dexterous manipulation. 

\textbf{Optimizing coating material}. After modifying the sensing surface shape, we try to obtain the best optical coating material for the sensing surface, while keeping the light locations fixed. We leverage the inverse design process to optimize the material. In both cases, we plot the evaluation criteria.

For the cylindrical sensing surface, we find that coating with specularity around 0.2 gives the best evaluation score, as shown in \Cref{fig:miniResults}Bii first row.  

For spherical sensing surface, we find that coating with specularity up to 0.4 gives the best evaluation score, as shown in \Cref{fig:miniResults}Bii second row. 

In both cases, we identify the specific material for the curved sensing shapes and are able to obtain sensing performance similar to the flat sensing surface design, which was already optimized using traditional manual design methods.

\subsection{Case study II: Rapid design of GelBelt}
\label{sec:rollerDesignExp}
\begin{figure*}[h]
    \centering
    \includegraphics[width=\linewidth]{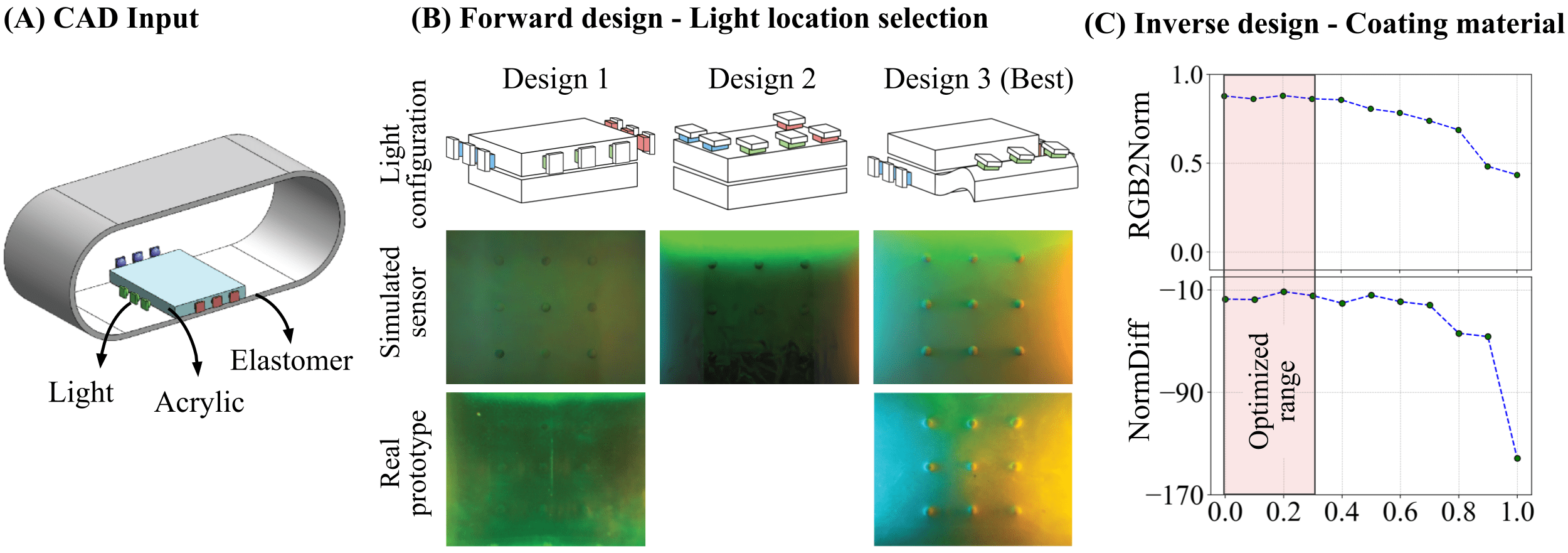}
    \caption{\textbf{Designing a new GelSight sensor, GelBelt}: We start with CAD design in (\textbf{A}) and create an optical design in OptiSense Studio. In (\textbf{B}), we perform forward design for light location selection. A human manually places lights at three plausible locations and uses simulation-driven evaluation criteria to select the best light configuration. It is evident that the perception of spherical indenters improves significantly with this approach. We also compare the images generated using the physical prototypes of GelBelt with Light Design 1 and Light Design 3. The simulated and real tactile images are shown in the middle and bottom rows. We see a close match in the tactile images and the superiority of Light Design 3 in real and simulated images.
    In (\textbf{C}), after selecting the best light configuration, we optimize the coating material using the inverse design procedure.}
    \label{fig:rollerResults}
\end{figure*}

The roller version of GelSight was first introduced in \cite{cao2023touchroller} to allow rapid perception of large surfaces. The authors created a cylindrical casing and molded elastomer over the casing by repositioning the sensor by just rolling; the sensor reduced the perception time while capturing consecutive images. 
To improve the accuracy of the roller sensor, we propose a new roller sensor design that contains a flat sensing surface between two rollers and uses OptiSense Studio to optimize the sensor design ~\citep{mirzaee2025gelbelt}. 
The design concept is visualized in \Cref{fig:rollerResults}A. The new design decouples the sensing surface design with the rolling phenomenon so that we can extend the sensing area without a direct impact on the wheel size. 

We show the forward design and inverse process using our design tool in \Cref{fig:rollerResults}. We consider variations in the choice of coating material, light location, and light types. The simulation time for GelBelt tactile image is \qty{8.9}{\s}. 

\textbf{Forward design process with light type and light locations.} 

Using the simulation toolbox, we investigate several configurations of the light locations and LED type to get the best design for the sensor, as shown in \Cref{fig:rollerResults}B. Initially,  it is aimed to have the LEDs illuminating on the side of the acrylic to mimic the GelSight mini light configuration. However, both in simulation and experiments it is shown that light poorly reaches the sensing surface. This is because, unlike GelSight sensors, the silicone is not cured or attached on top of the acrylic, letting a thin layer of air be trapped in between, resulting in the total internal reflection of the light in acrylic. To address this problem, we reconfigure the illumination system and rerun the simulation. As shown in \Cref{fig:rollerResults}B, lights are placed at different locations. The best configuration for the blue and red lights is to have them on the sides of the sensing surface but illuminated through silicone instead of acrylic. This configuration cannot be used for the green light, which limits its location to somewhere next to the acrylic. Several configurations are tested for green light. It was observed that having the green light illuminating at an angle through the silicone showed acceptable results. Then, in an improved design, the silicone is bent over a small roller to better guide light through the sensing surface. This way, the green light can travel further on the sensing surface.

\begin{figure}
    \centering
    \includegraphics[width=\columnwidth]{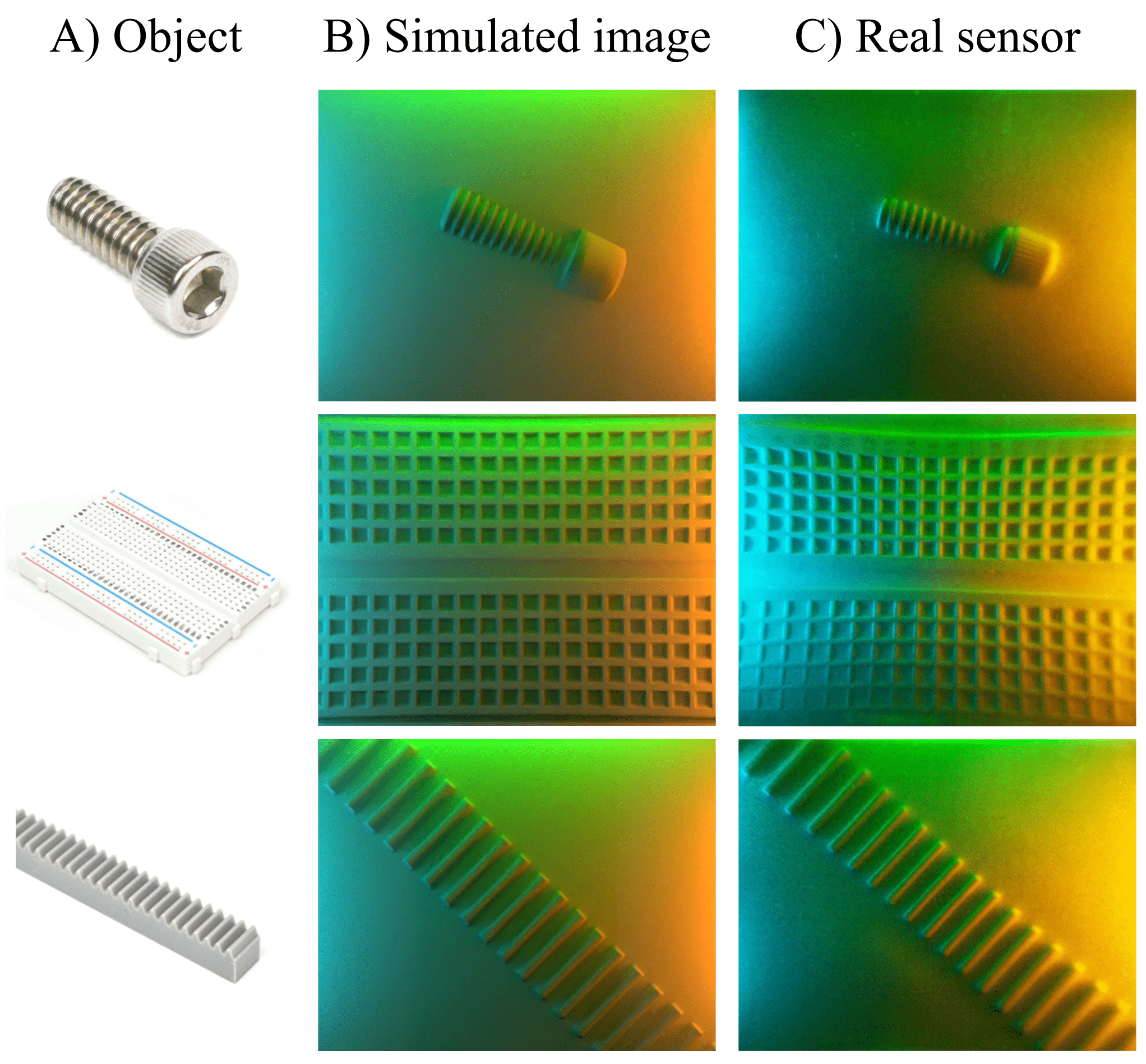}
    \caption{
    The simulated and real output of the optimized GelBelt sensor when contacting a screw, a breadboard, and a rack. 
    It is observed that the real sensor performance highly matches the simulation and well shows the object geometries.}
    \label{fig:simvsexp}
\end{figure}

\textbf{Inverse design of sensing surface coating material}. After optimizing the light location, we try to obtain the best optical coating material for the sensing surface, while keeping light locations fixed. We leverage the inverse design process to optimize material. We plot the normalized evaluation criteria for the best assessment. The results are visualized in \Cref{fig:rollerResults}C. We identify specularity less than 0.3 to be the best design using this process. 

\textbf{Real-world prototype of optimal GelBelt sensor}.
To verify the results of the simulation, we manufactured a prototype based on the optimal design. \Cref{fig:simvsexp} compares the output images of the proposed real-world sensor with those of the simulation when contacting different objects. It is observed that the simulation output of the approximate geometry is highly consistent with the images of the real sensor in all cases. This similarity strongly supports the validity of our design framework. However, researchers and designers can benefit from using OptiSense Studio and its modules to predict outcomes and optimize their design before fabricating the real-world sensor. 

\subsection{Case study III: Optical component shape variation and light variation for GelSight360}
\label{sec:G360_}
In this section, we consider the GelSight360 sensor, which was introduced in \cite{tippur2023gelsight360}. The authors used \textit{light-piping} and embedded lights to create a VBTS tactile sensor that could provide sensing in the forward direction without any occlusion. In this sensor, illumination design requires figuring out the light color setting and surface shape of multiple optical components for best perception. This makes the design problem particularly challenging. We first discuss light color variation and then discuss optical component shape variation using our objective functions. The simulation time for the GelSight360 tactile image is \qty{8.7}{\s}. 

\begin{figure*}
    \centering
    \includegraphics[width=1.0\textwidth]{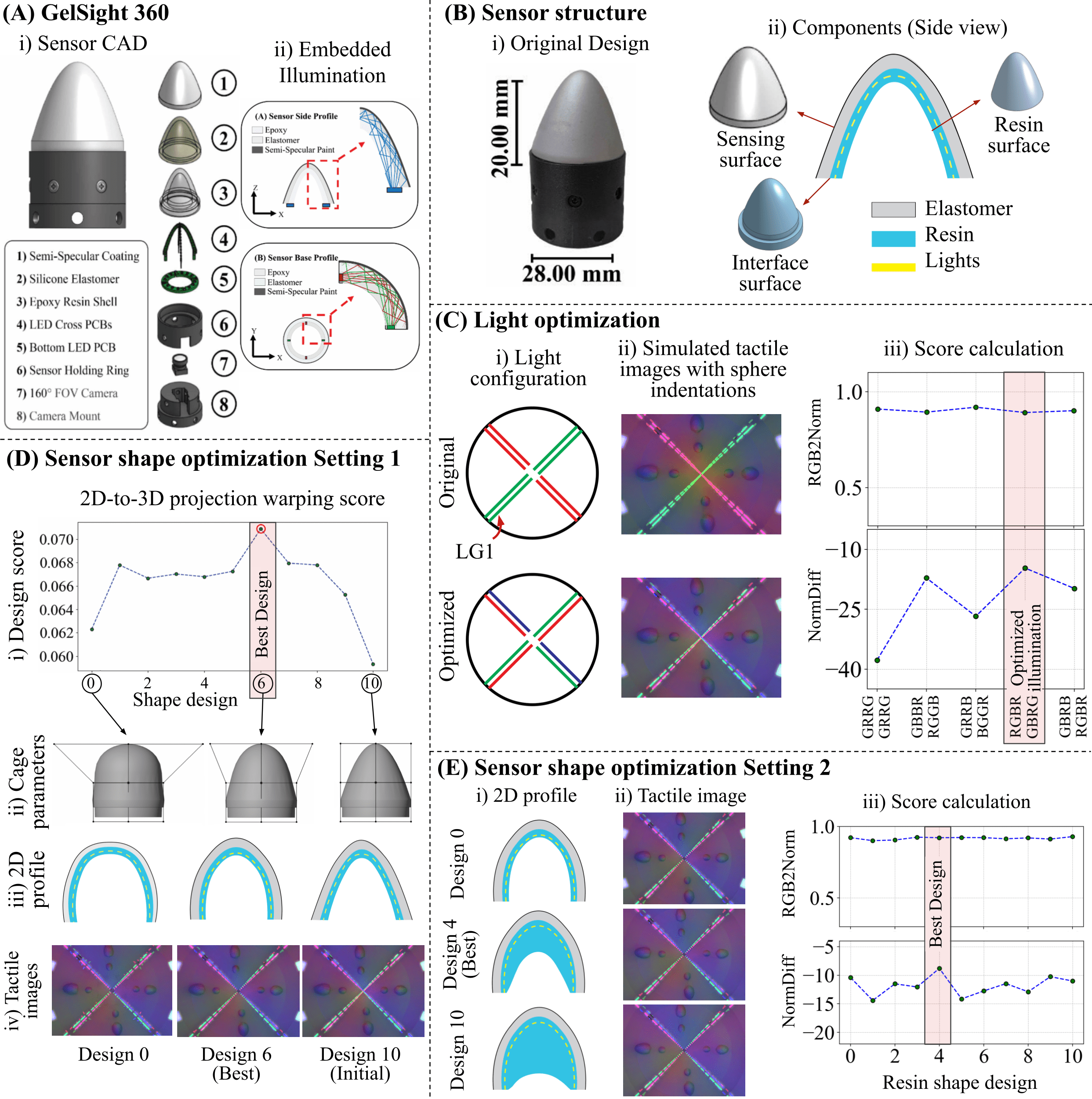}
    \caption{\textbf{GelSight360 shape and light variation description}: \textbf{(A)} Original GelSight360 sensor as introduced in \cite{tippur2023gelsight360}. 
    \textbf{(B)} shows the exploded view with labels and the corresponding side view with key components marked.     \textbf{(C)} (Ci) Top view is shown with the first Light Group (LG) shown as LG 1. We number the LG anti-clockwise starting from LG1. (Cii) shows the rendered tactile images for the original sensor and the best illumination setting based on the score in (Ciii). \textbf{(D)} shows the sensor shape design using the cage-based representations. (Di) The graph shows the 2D-to-3D projection warping score from the flatter sensor tip (Design 0) to the conical sensor tip (Design 10). The best design is shown in Design 6. \textbf{(E)} shows the inner-most resin shape design using cage-based representations. (Eiii) The graph shows the RGB2Norm and NormDiff scores of the thinnest resin shell (blue region) to the thickest resin shell (blue region). The best resin shell is Design 4.
    }
    \label{fig:gelsight360_desc}
\end{figure*}

\textbf{Light type variation.} In this experiment, we consider various light configuration with \textit{RGB2Norm} and \textit{NormDiff} objective functions. The vertical lights in the sensor are divided into 8 light groups, as shown in \Cref{fig:gelsight360_desc}C. \Cref{tab:gelsight360_light} shows the objective function values. According to the objective functions, \textit{RGBRGBRG} has the highest robustness to camera noise among similar \textit{RGB2Norm} scores for all designs considered. Note that performing the light variation in the real world requires designing and manufacturing new LED boards and embedding them in the resin shell, which is time-consuming and expensive for designers.

\begin{table}
    \centering
    \caption{\textbf{GelSight360 light type variation}: "R," "G," and "B" stand for red, green, and blue light colors, respectively. This table notes the score of the sensor designs with different illumination settings. Higher scores are better. }
    \begin{tabular}{||c|c|c||}
    \toprule
         Light configuration & \textit{NormDiff} $\uparrow$ & \textit{RGB2Normal} $\uparrow$ \\
         \toprule
         GRRGGRRG (original)	&	-37.745	&	0.909 \\
         GBBRRGGB	&	-17.172	&	0.893 \\
         GRRBBGGR	&	-26.769 &	\textbf{0.919}	\\
         GBRBRGBR & -19.810 & 0.901\\
         \textbf{RGBRGBRG}	&	\textbf{-14.696} & 0.891 \\
         \bottomrule
    \end{tabular}
    \label{tab:gelsight360_light}
\end{table}

\textbf{Optical component shape variation.} We optimize shape of optical components in two steps to obtain the best sensor. Firstly, we optimize the whole sensing surface shape using 2D-to-3D projection warping objective function, \Cref{fig:gelsight360_desc}D. In this step, we consider shape variations from a flat sensing tip (Design 0) to the conical sensing tip (Design 10). Secondly, we optimize the innermost resin surface to increase optical performance using NormDiff objective function. In this step, we consider resin shell variation from uniform thickness to almost flat inner-most resin surface, as shown in \Cref{fig:gelsight360_desc}E. 

For sensor shape setting 1, we obtain design 6 as the best design with 2D-to-3D projection warping score = \qty{0.071}. The final shape, corresponding 2D profiles, and tactile images are shown in \Cref{fig:gelsight360_desc}D (ii to iv). 

For sensor shape setting 2, we obtain design 4 as the best design with NormDiff score = \qty{-8.81}. 2D profiles and tactile images are shown in \Cref{fig:gelsight360_desc}E (i and ii).

\subsection{Case study IV: Sensor shape optimization for GelSight Svelte}
\label{sec:svelte_}
\begin{figure}
    \centering
    \includegraphics[width=\columnwidth]{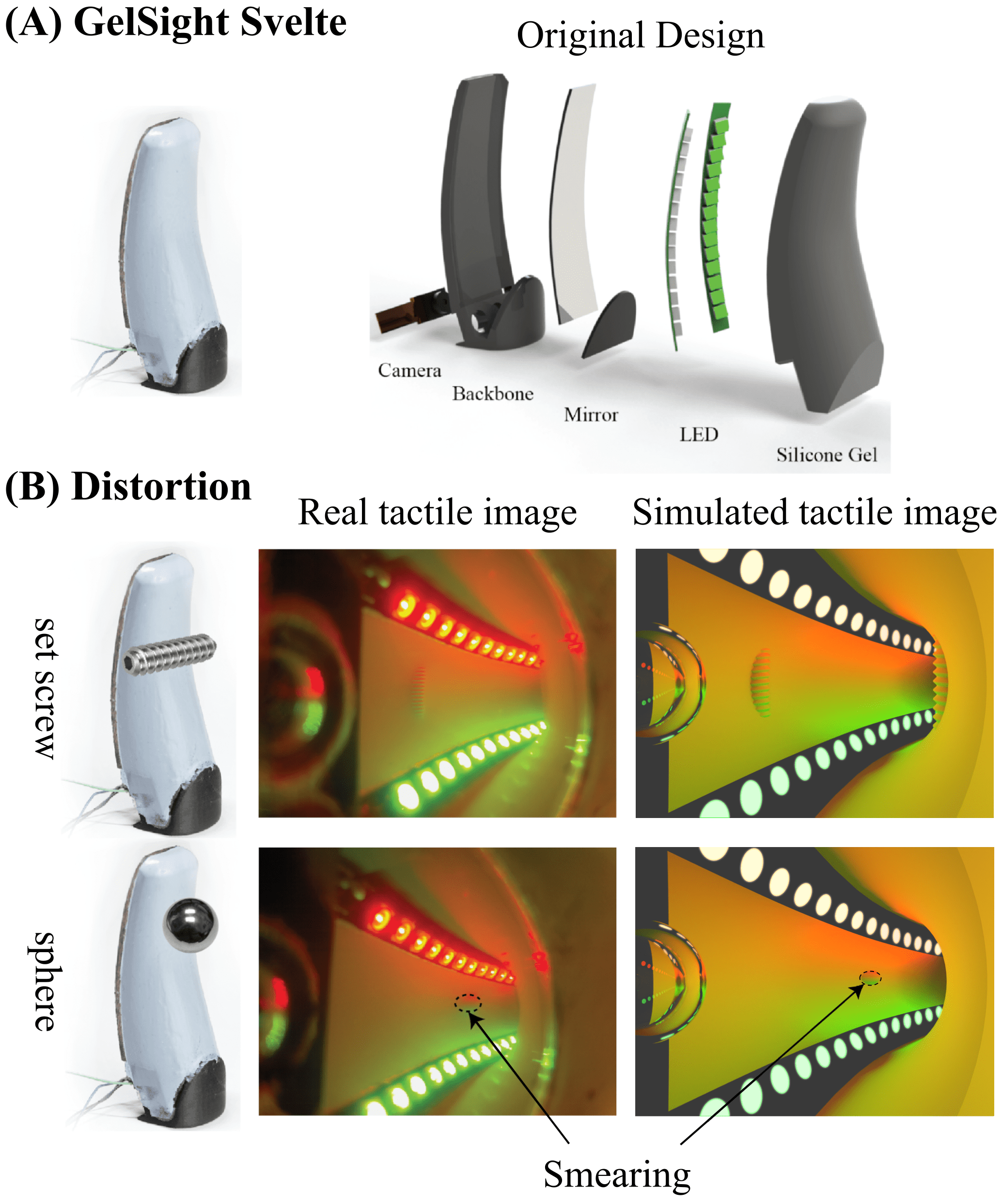}
    \caption{\textbf{GelSight Svelte issues}: \textbf{(A)} The original GelSight Svelte sensor was introduced in \cite{sveltezhao2023gelsight}.
    \textbf{(B)} highlights an issue in  We compare the simulated image against the real-world prototype tactile images. The simulated images are a close match to the real images. The top and bottom rows show images with setscrew and ideal sphere indenters at different sensing surface locations. As can be seen from the bottom row, the distortion depends on the indentation location. In the bottom row, the ideal sphere is "smeared" or distorted substantially, and is hard to perceive physical shape properties like sphere radius.}
    \label{fig:thin_image_comp}
\end{figure}

In this section, we consider the GelSight Svelte sensor, which was introduced in \cite{sveltezhao2023gelsight}. The authors used multiple mirrors to route the camera view to the full human-finger-shaped sensing surface. This allows sensing along the entire finger instead of just the tip.
The optics of the sensor were selected using 2D raytracing simulations. 
During our investigation, we noticed the ``smearing'' issue when the indenters are pressed on the sensing surface, as shown in \Cref{fig:thin_image_comp}. The amount of distortion depends on the indenting location on the sensing surface. This effect is caused by the larger back mirror design. The original design only considered sensing surface coverage in the 2D side view. 

We consider the design of the back mirror surface to alleviate this issue and improve sensing performance. We use \textit{As-orthographic-as-possible} (AOAP) objective function in this experiment. To show a proof of concept, we first consider a simplified sensing surface and focus on improving perception at the center of the sensing surface. The key optical surfaces are shown in \Cref{fig:shape_optim}A. We consider cage-based parameterization of the larger mirror surface, M1. This reduces the search space by orders of magnitude from 22680 to 81. We initialize the cage using the original mirror shape. We choose optimization parameters $\mathcal{C}_{\text{min}}$ such that M1 is flat and $\mathcal{C}_{\text{max}}$ such that M1 has the largest curvature possible without intersecting with the sensing surface. We use CMA-ES~\citep{hansen2016cma} to optimize the shape parameters. 

\begin{figure}
    \centering
    \includegraphics[width=\columnwidth]{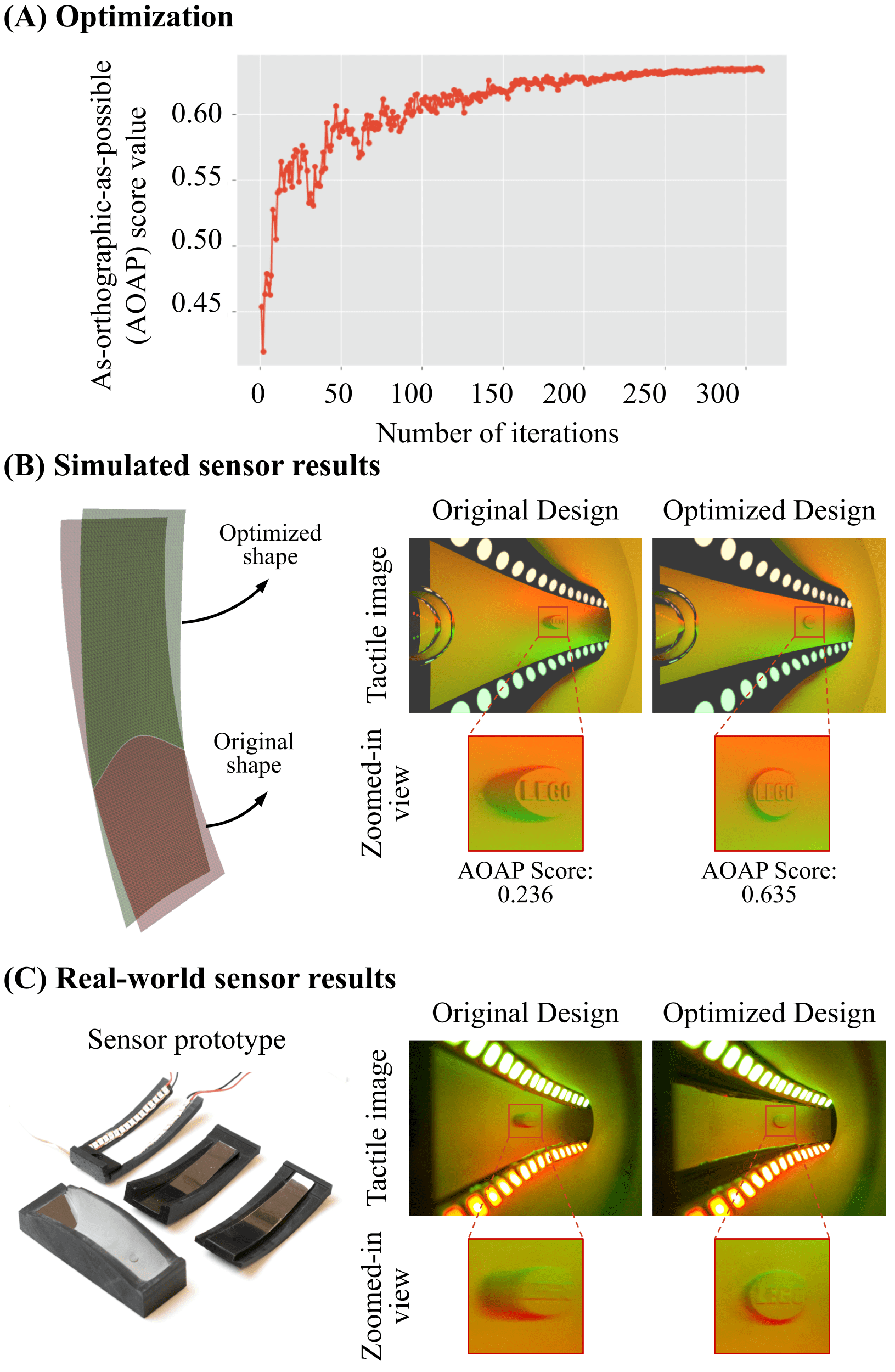}
    \caption{\textbf{GelSight Svelte shape optimization results}: We show the shape optimization results for the GelSight Svelte sensor. \textbf{(A)} shows the AOAP function score during the CMAES optimization procedure. \textbf{(B)} shows the initial and optimized larger mirror surface mesh in red and green, respectively. We also simulated tactile images for the two designs. The optimized design has significantly reduced distortion as compared to the initial design. \textbf{(C)} We manufactured sensor prototypes to compare the improvement in the real world for initial and optimized design. The left visual shows our prototype. We show the tactile images from the real-world prototypes and their zoomed-in view. The real and simulated images of the initial design both show significant distortion of the Lego block. The issue is resolved completely in real and simulated images for optimized design. }
    \label{fig:thin_shape_optim_res}
\end{figure}

The optimization curve is shown in \Cref{fig:thin_shape_optim_res}B left. The AOAP score for the initial and optimized design is \num{0.236} and \num{0.635}. 
As can be seen from the rendered tactile images in \Cref{fig:thin_shape_optim_res}B right, the "smearing" effect or distortion is almost gone in the optimized design. Thus, our shape optimization pipeline could be used to obtain the best optical component shapes to reduce optical distortion and improve shape perception. Note that this approach can be applied to any optical surface design. 

\section{Discussion and conclusion}
In this paper, we presented a design framework for the development of a variety of GelSight-like sensors and created an interactive interface for rapid sensor design and optimization. We demonstrated the application of the design framework by rapidly optimizing the existing GelSight-family tactile sensors and creating new sensors in simulation. We then prototyped different designs to compare the simulation and real-world results. Our design framework codifies the design process for vision-based tactile sensors into design modules and leverages the digital design process with optical simulation to optimize sensor design. In addition, we provide a component library of commonly used optical components in vision-based tactile sensors to aid in the design process.

\subsection{Why Choosing Those Objective Functions?}
The GelSight sensors are primarily designed to precisely measure the 3D geometry of contact surfaces, and their application in various robotic perception tasks depends on this capability. Our objective functions quantitatively assess a GelSight sensor's performance in accurately capturing surface geometry, independent of shape reconstruction algorithms or signal processing techniques. Sensors that achieve high scores with our objective functions can provide high-quality geometric measurement signals across various tasks, leading to improved perception outcomes.

We demonstrate this through two simulated tasks: estimating the misalignment angle between a tactile sensor and a plane, and determining the radius of a contacting cylinder. For both tasks, we compiled a small tactile dataset and employed simple random forest regression models to generate results. Random forests were chosen for their robustness and generalizability, even with limited data.

\begin{figure*}
    \centering
    \includegraphics[width=1.0\textwidth]{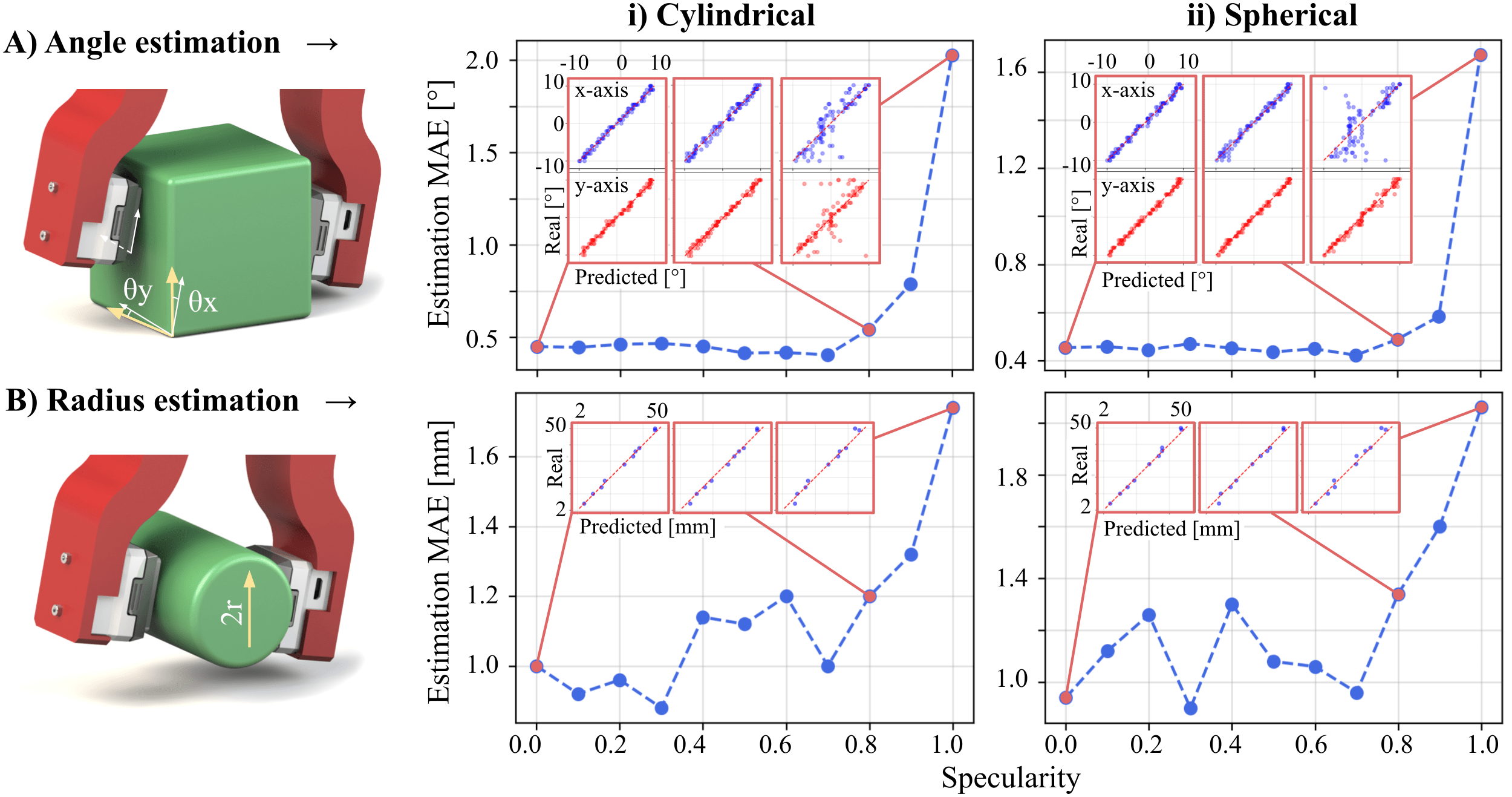}
    \caption{\textbf{The effect of specularity on the GelSight Mini performance in tasks that do not directly concern geometry reconstruction:} We train a simple random forest model to directly estimate \textbf{(A)} the misalignment angle between the sensor and surface in contact and \textbf{(B)} surface curvature from RGB information in the image. The first column illustrates the scenario setting. \textbf{i} and \textbf{ii} show the results based on data collected with different sensors in simulation, and the results show a similar trend as the RGB2Norm and NormDiff objective functions. This shows that the optimized sensor design based on our objective functions will help real robotic tasks even when using machine learning methods for data processing.
    }
    \label{fig:perception}
\end{figure*}

\textbf{Estimating the misalignment angle with a plane.} 
Accurately estimating the misalignment angle between a tactile sensor and a target plane is crucial for ensuring optimal contact. This estimation enables a robot to adjust its movements, reducing the misalignment and achieving full contact. To estimate this angle using GelSight measurements, we simulated a GelSight Mini sensor with both cylindrical and spherical surfaces contacting a virtual plane at various angles along two axes, as depicted in \Cref{fig:perception}A. The sensor setup mirrors the experiment detailed in Section \ref{chpt:MiniExperiment}. We extracted the mean RGB values from a cropped central region of the tactile images as the primary feature and utilized a random forest regression model to predict the misalignment angle. The results, shown in \Cref{fig:perception}A, indicate that our optimized sensor achieves a mean absolute error (MAE) as low as 0.5$^{\circ}$ when coated with low-specularity materials. As the material's specularity increases, the MAE also rises. This trend aligns with the RGB2Norm and NormDiff scores of the sensors, suggesting that our objective functions can effectively predict the sensor's performance in contact angle estimation.

\textbf{Estimating the surface curvature size.} We set up the experiment as using GelSight Mini sensors to contact cylinders with varying radii in simulation, as shown in \Cref{fig:perception}B. 
We calculate the standard deviation within the contact area of each color channel and use them as inputs to train a random forest regression model. We do this process for both cylindrical and spherical gelpads, with results illustrated in  \Cref{fig:perception}B. In both cases, we observe a similar trend as we had in the optimization stage, which supports the applicability of our objective function for tasks more than geometry measurement.

\textbf{Signal-to-noise ratio.}
To evaluate the robustness and reliability of data collected from different sensor designs, we employ a model-free signal-to-noise ratio (SNR). This SNR formulation provides a quantitative measure of how well the sensor data is structured and whether the target variables exhibit consistency within local regions of the feature space. In traditional signal processing, SNR is a fundamental metric for assessing data quality, where a higher SNR indicates a stronger signal relative to noise. In the context of regression-based sensor evaluation, we redefine this concept by considering the global variance of the target variable as the signal and the average local variance among k-nearest neighbors as the noise, Equation~\eqref{eq:snr}. The k-nearest neighbors are calculated in the feature space while variance is calculated in the prediction space. This approach allows us to determine whether variations in sensor measurements are structured and predictable (high SNR) or dominated by local inconsistencies and noise (low SNR).

\begin{equation}
\text{SNR} = \frac{\sigma^2_{\text{global}}}{\mathbb{E}[\sigma^2_{\text{local}}]}
\label{eq:snr}
\end{equation}

\begin{figure}
    \centering
    \includegraphics[width=\linewidth]{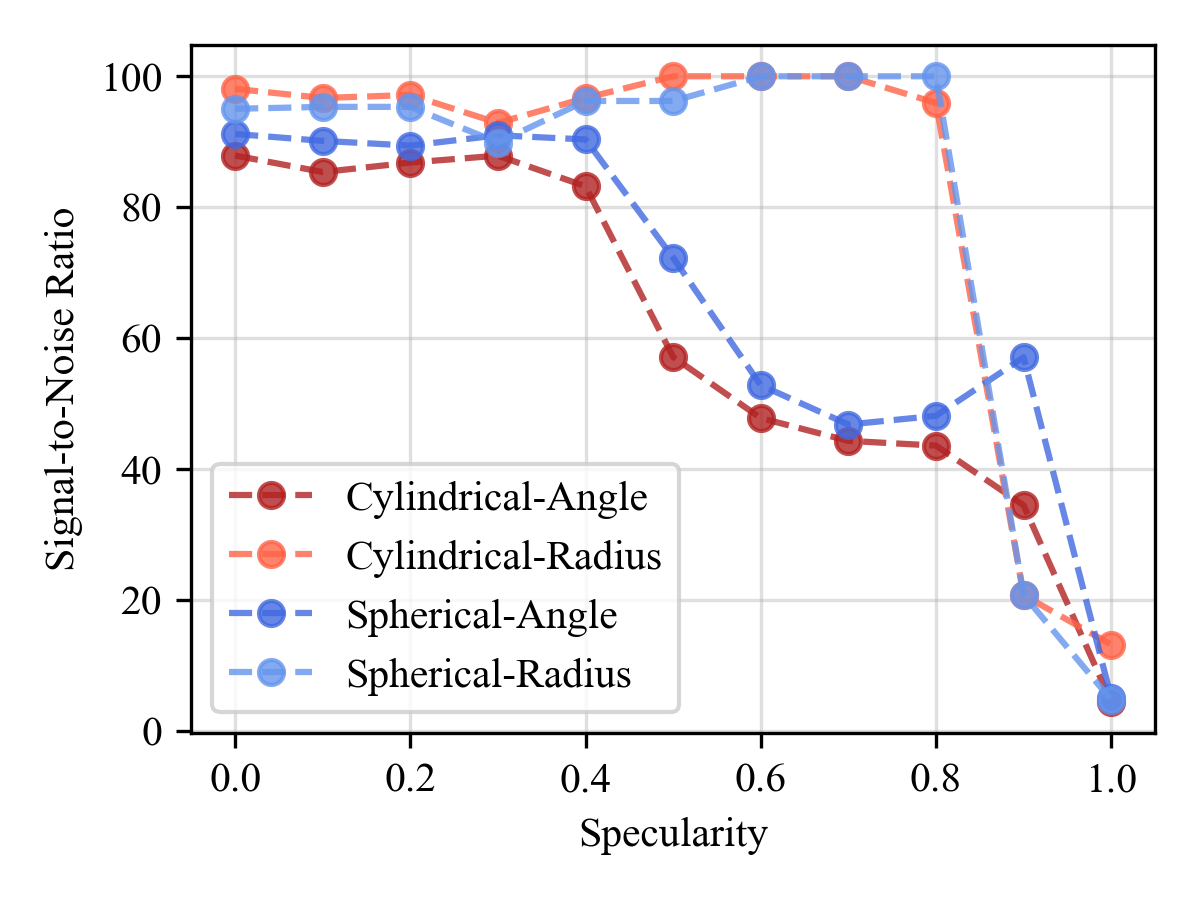}
    \caption{Singnal-to-nise ratio of sensors with varying specularity and shape. We define the signal-to-noise ratio as the ratio between the global variance and local variance of k-nearest neighbors. We see a similar trend of decreasing scores as the specularity increases, which is similar to the RGB2Norm, NormDiff, and the previous experimental results, especially for the angle misalignment estimation.}
    \label{fig:snr}
\end{figure}

\Cref{fig:snr} illustrates the SNR values of the collected data for previous experiments on angle misalignment and radius estimation. We observe that the angle estimation results are highly similar to the result of RGB2Norm and NormDiff objective functions for sensors with varying specularity. The radius estimation trend, similar to the random forest result, shows the decreasing performance with high specularity that was observed in the sensor optimization section, suggesting that our objective functions can adequately estimate the sensor's performance in tasks more than geometry measurement.

\subsection{Future Work: More Precise Modeling and Optimization}

A major limitation of our current framework lies in the lack of precise modeling of sensor artifacts. While our current objective functions describe the most important metrics for the design of GelSight sensors, but some corner cases or specific design goals are not considered. Some design goal examples are the sensor noise caused by light shadows, which commonly appear for some sensors like DIGIT~\citep{lambeta2020digit}, and detecting high-frequency signals such as surface fractures instead of precise shape measurement. Our design interface, OptiSense Studio, allows users to incorporate user-defined objective functions for these specific goals. Developing objective functions for different design goals is left for future work.

Another commonly seen artifact comes from manufacturing error, which causes some gap between the simulated sensor design and real prototypes. We believe that with the existing manufacturing error and the sim-to-real gap, our framework provides a way to obtain the likely best design. In the future, we will further explore how to model the manufacturing variance in our design framework. 

Our current framework does not consider the choice of sensor material due to mechanical properties such as elasticity. This is because they play a relatively minor role in sensor performance and are generally straightforward to optimize. The choice of sensor material's mechanical properties is independent of the optical performance and is highly application-driven. In other words, for sensors with the same optical design, replacing one elastomer with another of different elasticity will not affect the sensor’s capability for shape measurement. Instead, users can directly choose materials that exhibit proper amount of deformation under the targeted force measurement range. 

\subsection{Future Work: Task-driven Optimization}

The proposed optimization process aims to enhance the overall quality of shape measurement in GelSight sensors, thereby fundamentally improving their performance across various tactile perception tasks. However, developing objective functions inherently involves dimensionality reduction of sensor measurements, introducing a certain ``bias.'' For specific tasks, objective functions with different biases may be more advantageous.

Our current framework, OptiSense Studio, supports user-defined objective functions, enabling users to tailor functions to their specific tasks. In future work, we plan to explore more effective methods for designing task-driven objective functions. This includes defining new functions suited to particular categories of tactile perceptual tasks and integrating these tasks directly into the optimization process.

Additionally, our existing design framework focuses solely on optimizing the sensor for improved measurement capabilities, without considering function-driven shape design. For example, while both spherical and cylindrical gel pads for the GelSight Mini exhibit similar optical performances, the spherical sensor better facilitates object pivoting on its surface. Our future research will investigate methods to evaluate sensor performance in specific control or manipulation tasks and incorporate these considerations into the optimization of the sensor's optical system.

\subsection{Future Work: Optimizing Designs for Other Vision-based Tactile Sensors}

We show that the pipeline performed well for the regular designs and simulation of the GelSight sensor family, and we expect that our approach can be extended to incorporate non-regular optical components for new vision-based sensor designs. For example, we hope to incorporate an approximate model of fluorescent lighting as introduced in \cite{finrayliu2022gelsight} to allow the design of the FinRay GelSight family.
We also look forward to incorporating absorption surface materials and translucent materials to simulate and optimize sensors like 9DTact~\citep{lin20239dtact}.

To extend our method to other types of vision-based tactile sensors, we will need to design new evaluation functions based on the working principles of the sensors. For example, ~\cite{lin20239dtact} relies on a single color intensity to reconstruct the geometry. In this case, users can use the general idea---to create a mapping between the measured image signal and tactile signal---of the evaluation criteria proposed in this work. 

Our framework does not simulate and optimize the mechanical properties of the sensors. For the GelSight family or other similar optical sensors that aim at measuring the geometry, this is a less important design factor.
However, for sensors that use visual information to indicate mechanical signals, such as the TacTip family~\citep{ward2018tactip} that uses marker displacement to measure the contact shape and forces, our current framework does not work. This is because our current framework assumes a thin, conformable contact surface rather than a mechanically complex skin deformation. Our future work will incorporate mechanical simulation of the sensors to extend the design space. We will also incoperate the mareker distribution pattern as the a design factor for those sensors. 

\bibliographystyle{SageH}
\bibliography{ref}

\clearpage

\section{Component library}\label{app:componentsLib}
\begin{figure*}
    \includegraphics[width=\linewidth]{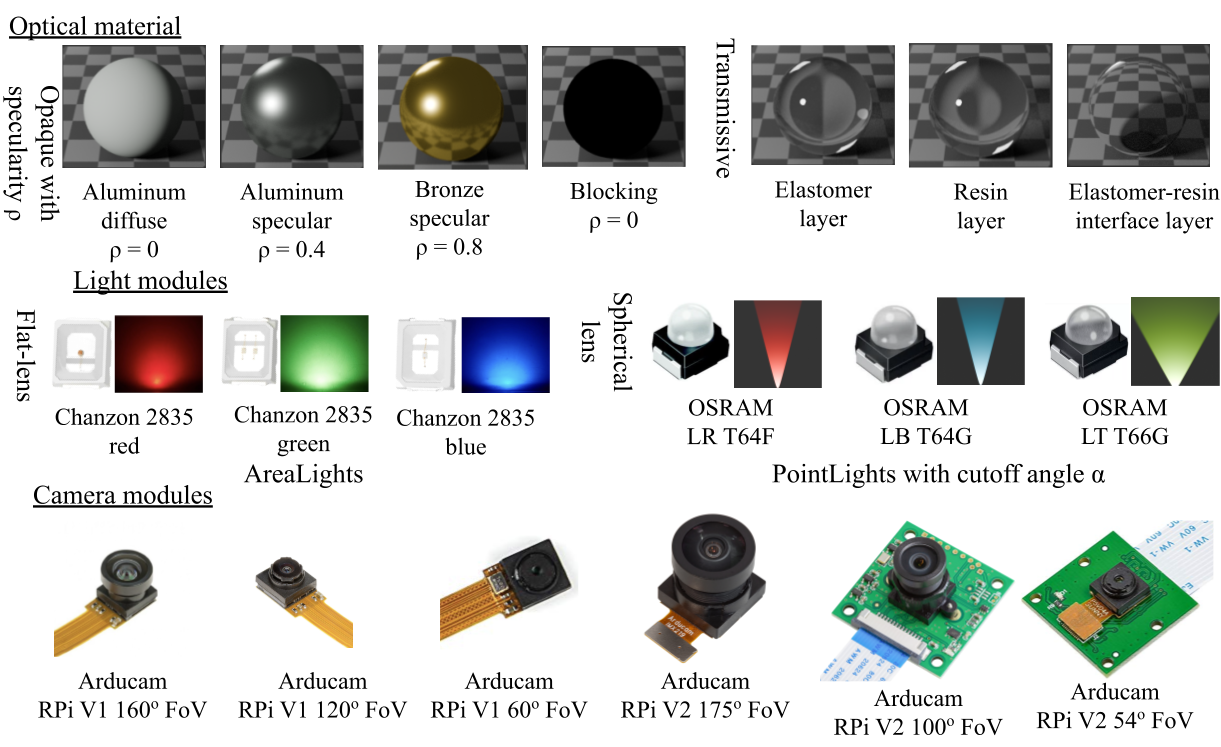}
    \caption{\textbf{Component library}: This figure shows the various components present in the library provided with our design interface. These components cover the design space of the GelSight sensor family and provide relevant design spaces to develop new sensors.}
    \label{fig:componentLib}
\end{figure*}

We provide 7 optical materials, obtained 6 light models and 6 camera types as shown in \Cref{fig:componentLib}. 

\section{Modeling the sensor}
\label{sec:pipeline}

\begin{figure*}
    \includegraphics[width=\textwidth]{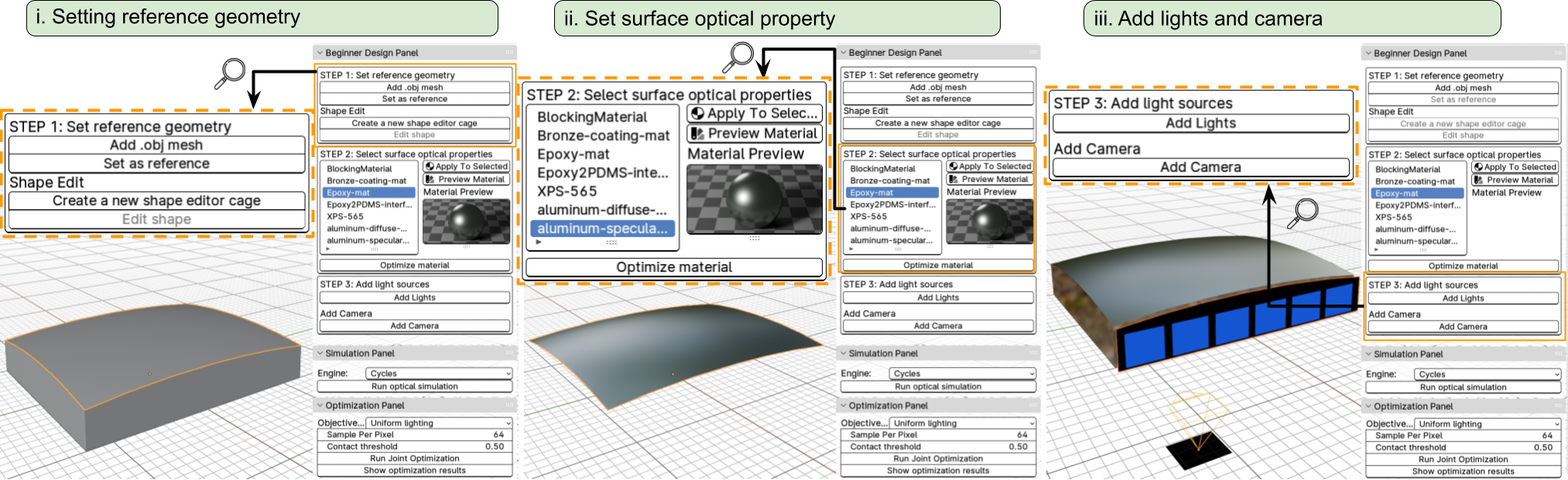}
    \caption{\textbf{Digital design guideline}: the three steps of the interactive design pipeline. i) Importing CAD shapes and setting them as reference geometries for optical elements; ii) Assigning material properties to the component from the component library or using user-defined materials; iii) Adding lights and the camera using reference geometries. The lights and the camera are chosen from the component library. }
    \label{fig:gui}
\end{figure*}

We modularize the design procedure into three steps and provide a library of optical components to aid in design. The user workflow consists of the following steps (see \Cref{fig:gui}).

The user starts with an idea for the sensor design or starts from a previous design. The user provides sensing surface geometries, initial light location, and camera location to the \textit{OptiSense Studio}. Thereafter, the users follow the given steps to generate an optical sensor design which can be simulated and perform design optimization in our software. For a brief tutorial, please refer to the Appendix \Cref{app:guiTutorial}

\textbf{Step 1: Setting reference geometry.} Users can create sensors of arbitrary geometry by importing shape reference geometry as \textit{.obj} surface meshes. We found that users prefer to choose their favorite CAD tools (Solidworks, Autodesk Fusion 360, and OnShape) for shape design. Specifically, users need to select \textit{Sensing surface reference}, \textit{Camera reference}, \textit{Light reference}, and \textit{Support structure reference}. Users can also select multiple surfaces for reference. Additionally, users can select \textit{Indenter reference}, \textit{Mirror reference}, \textit{Blocking reference},   \textit{Interface reference}, and multiple light references as optical elements. 

\textbf{Step 2: Selecting optical properties for surfaces.} After the addition of key surfaces, the user needs to assign optical properties to each surface. Based on the literature review of camera-based tactile sensors we provide a library of optical materials, which support refraction with rough interfaces, reflection with rough interfaces, and blocking of light paths. Specifically, we provide diffuse coating material~\cite{slipdetection}, semi-specular coating material~\cite{brandenroundsensor}, PDMS refractive surface~\cite{slipdetection}, and Epoxy rough refractive surface~\cite{yuan2017gelsight}. For a detailed overview refer to the \Cref{app:componentsLib}. All the materials were obtained by performing optical experiments in the lab and fitting analytical models available in the physics-based rendering and material modeling literature.

\textbf{Step 3: Adding light sources and camera. } Next the user chooses the light reference surface from \textit{Reference} collection and selects the light type in the pop-up menu. Our library contains physically accurate light models based on the data available from the LED manufacturers. We provide point light sources with IES profiles sourced from manufacturers, spotlights with cut-off angles sourced from LED datasheets, and area light sources with dimensions sourced from LED datasheets. For a detailed overview refer to the \Cref{app:componentsLib}. This allows accurate simulation and accurate physical light placement such that the sensor design preview is physically accurate. 

For adding the camera, the process is similar to adding lights. Choose \textit{Camera reference} surface from the \textit{Reference} collection, click on \textbf{Add Camera} button, and select the desired camera. Our library provides commonly used Raspberry Pi cameras with field-of-view ranging from \ang{60} to \ang{160}. For a detailed overview refer to the \Cref{app:componentsLib}.

\section{Design tutorial in OptiSense studio}
\label{app:guiTutorial}
In this section, we give a short tutorial on how to set up a camera-based sensor design, using GelSight Mini as an example.

\textbf{Step 1: Adding shapes}

\noindent To add shapes click on the \textbf{Add .obj mesh} button and assign them as reference geometry by clicking on the \textbf{Set as reference} button. We provide a reference surface for each of our design module. Each module performs optical-component-specific shape and material initialization in the final optical system. 

Users can edit the surface shape using cage-based representation~\cite{differentiableRobotDesign21}. To create a cage around the shape, select the shape and click the \textbf{Create a new cage} button. To edit the shape, click the \textbf{Edit shape} button, select the cage vertices to move, and then move the vertices to change the surface shape. As noted in \cite{differentiableRobotDesign21}, the cage-based representation is differentiable and in the future is amenable to differentiable sensor shape design. 

\textbf{Step 2: Assign optical material property}

\noindent To assign the optical material, the user selects the desired optical surface from the \textit{OpticalSystem} collection, selects the desired material from the library and then clicks on \textbf{Apply To Selected} button. The user can also preview the material before assigning it by selecting the material and clicking the \textbf{Preview Material} button. 

\textbf{Step 3: Add light and camera}

\noindent For adding the lights, choose \textit{Light reference} surface from the \textit{Reference} collection, click on \textbf{Add Light} button, and select the desired light. Our library provides commonly used LEDs with flat and spherical lenses, commonly used in camera-based sensors. For a detailed overview refer to the \Cref{app:componentsLib}.

For adding the camera, the process is similar to adding lights. Choose \textit{Camera reference} surface from the \textit{Reference} collection, click on \textbf{Add Camera} button, and select the desired camera. Our library provides commonly used Raspberry Pi cameras with a field of view ranging from \ang{60} to \ang{160}. For a detailed overview, refer to \Cref{app:componentsLib}.

\section{Fabrication of the new GelBelt Sensor}
\noindent After obtaining the best design for GelBelt using our digital design framework, we manufactured a real-world prototype and tested its feasibility. PLA was used to 3D print the sensor frame and handles. The wheels were 3D printed by Form 3+ printer in Black Resin Material to have a smoother print surface. The Acrylic part of the sensor was laser cut to the shape and fixed in its housing using thin double-sided tape. The belt is made of Silicone XP565 (Silicone Inc.) while coated with Aluminum powder on the contact surface. Silicone itself cannot slide on the acrylic because of the high frictional force between two surfaces; therefore, an intermediate layer is required to complete the task of sliding. For this purpose, wide clear tape was attached to the inner side of the belt as it showed acceptable adhesion to silicone on the glue side while having a small friction with acrylic on the other side. To make the prototype of the sensor, the belt was fabricated by having a flat mold. It should be mentioned that the belt could be fabricated using a circular mold to have continuous rolling over the surface, which will be considered in the future. After Silicone was cured and coated with aluminum powder, the belt was removed from the mold and a wide tape was attached to the uncoated side of it. To have the complete belt, the belt was bent all over the rollers and then attached using the wide tape.
Regarding the lights, several SMD 3528 LEDs were linearly arranged for each of the red, green, and blue lights. The lights were soldered on a PCB and then fixed on the sensor frame using screws.

\end{document}